\pdfoutput=1

\documentclass[10pt,twocolumn,letterpaper]{article}
\usepackage[pagenumbers]{iccv} 

\usepackage{graphicx}
\usepackage{booktabs}
\usepackage{graphicx}
\usepackage{booktabs}
\usepackage{multirow}
\usepackage{multicol}
\usepackage{amsmath}
\usepackage{amssymb}
\usepackage{pifont}
\usepackage{xcolor}
\usepackage{colortbl} 
\usepackage{array}
\usepackage{amsmath}
\usepackage{amssymb}
\usepackage{algpseudocode}
\usepackage{listings}
\usepackage{algorithm}
\usepackage{alltt,xcolor}
\usepackage{inconsolata}
\definecolor{lightblue}{rgb}{0.87, 0.94, 1}
\newcommand{\xmark}{{\ding{55}}}%
\newcommand{\garrow}[1]{\,\textcolor[HTML]{2DB83D}{\scalebox{0.75}{\scalebox{0.975}[1]{\;\textbf{$\uparrow$}\,\textbf{#1}\hspace{-21pt}}}}}
\newcommand{\greenarrow}[1]{\,\textcolor[HTML]{2DB83D}{\scalebox{0.75}{\scalebox{0.975}[1]{\,\textbf{$\uparrow$}\,\textbf{#1}\hspace{-23pt}}}}}
\newcommand{\redarrow}[1]{\,\textcolor[HTML]{E57373}{\scalebox{0.75}{\scalebox{0.975}[1]{\,\textbf{$\downarrow$}\,\textbf{#1}\hspace{-23pt}}}}}
\newcommand{\graytext}[1]{\,\textcolor[HTML]{808080}{\scalebox{0.75}{\scalebox{0.975}[1]{\;\,\textbf{#1}\hspace{-19pt}}}}}
\usepackage{arydshln}
\usepackage[subtle]{savetrees}

\definecolor{iccvblue}{rgb}{0.21,0.49,0.74}
\usepackage[pagebackref,breaklinks,colorlinks,allcolors=iccvblue]{hyperref}

\title{Shot-by-Shot: Film-Grammar-Aware Training-Free Audio Description Generation}

\author{
Junyu Xie $^{1}$ \quad Tengda Han$^{1}$ \quad Max Bain$^{1}$ \quad
Arsha Nagrani$^{1}$ \quad Eshika Khandelwal $^{2,3}$  \\ G\"ul Varol$^{1,3}$ \quad Weidi Xie$^{1,4}$ \quad Andrew Zisserman$^1$\\
{\small$^1$Visual Geometry Group, University of Oxford} \quad
{\small$^2$ CVIT, IIIT Hyderabad}
\\
{\small$^3$LIGM, \'Ecole des Ponts ParisTech}
\quad {\small$^4$SAI, Shanghai Jiao Tong University} \\
{\small\url{https://www.robots.ox.ac.uk/vgg/research/shot-by-shot/}}}

\begin{document}

\twocolumn[{%
\renewcommand\twocolumn[1][]{#1}%
\maketitle

\begin{center}
    \centering
    \vspace{-1.5mm}
    \captionsetup{type=figure,skip=0pt}
    \includegraphics[width=0.975\linewidth]{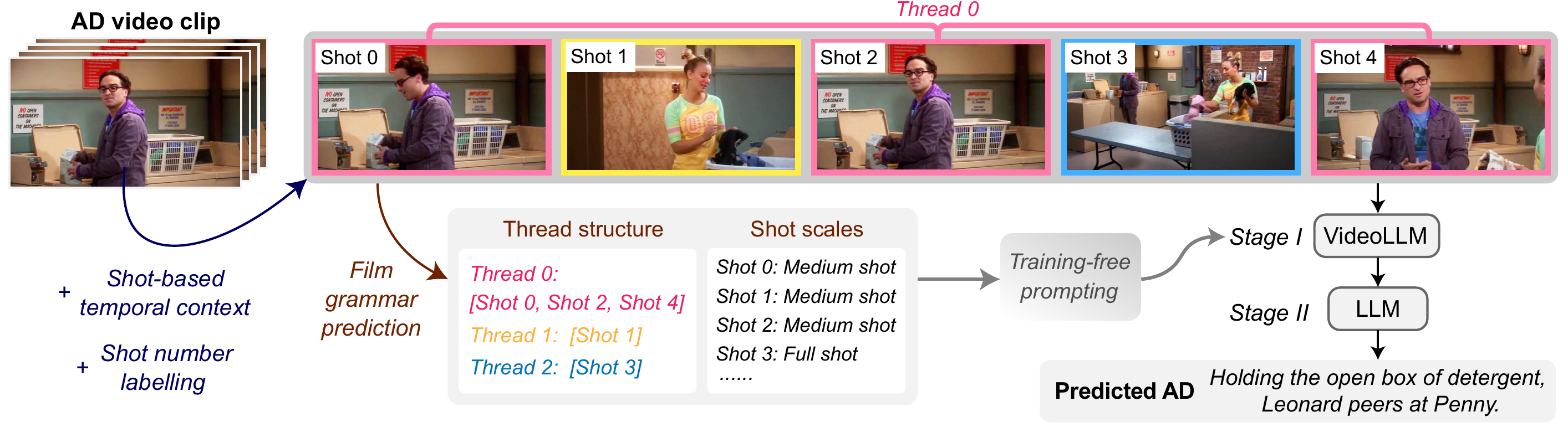}
    \vspace{1.5mm}
    \captionof{figure}{\small \textbf{Overview of our training-free Audio Description (AD) generation framework.} The input video clip (corresponding to the AD interval) is extended to include adjacent shots, providing richer temporal visual context. Each video frame is labelled with its corresponding shot number. The extended clip then undergoes a film grammar prediction process, where the thread structure and shot scales are estimated. The AD is generated in two stages: Stage I utilises the predicted cinematic information as training-free prompt guidance to produce dense video descriptions. Stage II then employs an LLM to generate a summarised AD output.}
    \label{fig:overview}
\end{center}
\vspace{0.15cm}
}]

\begin{abstract}
Our objective is the automatic generation of Audio Descriptions (ADs)
for edited video material, such as movies and TV series.
To achieve this, we propose a two-stage framework that leverages ``shots'' as the fundamental units of video understanding.  
This includes extending temporal context to neighbouring shots and incorporating film grammar devices, such as shot scales and thread structures, to guide AD generation.  
Our method is compatible with both open-source and proprietary Visual-Language Models (VLMs), integrating expert knowledge from add-on modules without requiring additional training of the VLMs.  
We achieve state-of-the-art performance among all prior training-free approaches and even surpass fine-tuned methods on several benchmarks. 
To evaluate the quality of predicted ADs, we introduce a new evaluation measure -- an action score -- specifically targeted to assessing this important aspect of AD.  Additionally, we propose a novel evaluation protocol that treats automatic frameworks as AD generation assistants and asks them to generate multiple candidate ADs for selection.
\end{abstract}    
\section{Introduction}
\vspace{-0.5mm}

\label{sec:intro}

In movies and TV series, Audio Descriptions (ADs) are narrations provided for the visually impaired, conveying visual information to complement the original soundtrack. 
Their purpose is to ensure a continuous and coherent narrative flow, enabling audiences to follow the plot effectively.
Unlike video captions, ADs are constrained by length, prioritising the most visually salient and story-centric information, such as character dynamics and significant objects, whilst omitting redundant details like background figures or unchanging locations. 
Additionally, ADs are typically produced by professional narrators in a specific style and format, ensuring coherence while not interfering with the original audio.

With the advent of Visual-Language Models~(VLMs), there has been a growth of interest in automatically generating ADs for both movies~\cite{ye-etal-2024-mmad-multi, Zhang_2024_CVPR, Han23, Han23a, Han24, uniad, movieseq, chu2024llmadlargelanguagemodel} and TV material~\cite{Xie24b, distinctad}.
However, as anyone who has ever watched a movie or read a book about film editing knows -- the fundamental unit of edited video material is the {\em shot}, not the frame~\cite{Monaco09, Katz19, Murch21}. Shots are used to {\em structure} the video material, defining the granularity and temporal context through choices in its scale (close-up, long shot, etc.) and its movement (panning, tracking, etc.).
The ``film grammar'' is then used to convey meaning through specific editing choices on shot transitions (cuts, fades, etc.), durations, and composition (threads, montage, etc.). 

Current approaches to automating AD generation, and video large language models (VideoLLMs) 
in general,  
are not shot-aware, hindering interpretation of edited video material that has
frequent shot transitions~\cite{tang2024vidcompostion}. In this paper, our principal objective is to incorporate shot information and editing structure into the AD generation process.
To this end, we consider temporal context in terms of shots, and take account of the two key properties of {\em thread structure} and {\em shot scale}. 

Thread structure identifies the sequence of shots captured with the same camera.
An example of a scene with multiple simultaneous threads in space-time is shown in~\cref{fig:thread}.  
We develop a robust thread clustering method and use its predictions to
guide VLMs in understanding shot-wise relationships.

Shot scale typically  implies the type of content of the frame. For instance, close-up shots often highlight characters' facial expressions or gaze interactions, while long shots tend to depict the overall environment and ambience of the scene, as shown in \cref{fig:shot_scale} (top).
We leverage this property by developing an off-the-shelf
shot scale classifier and employing a scale-dependent prompting strategy
to improve the contextual relevance of VLM-generated descriptions.

In this paper, we develop a {\em training-free} AD generation
approach. Inspired by AutoAD-Zero~\cite{Xie24b}, as illustrated in~\cref{fig:overview}, we adopt a two-stage pipeline in which a \mbox{VideoLLM} generates dense text descriptions in the first stage, followed by an LLM that produces the final AD outputs from this text. We improve the  effectiveness
of training-free approaches by incorporating two key improvements based on: (i)
shot-based temporal context; and (ii)
the shot scale and thread structure, that are both crucial to cinematic composition and understanding.
Previous methods could describe human-object interactions or human-human interactions, such as looks~\cite{Marin20}, within a frame or shot. With the improvements we introduce, the generated AD also includes such interactions when they are implied by the shot structure.

As well as the challenge of generating ADs, another challenge is {\em how to evaluate} the predicted ADs. Apart from conventional metrics~\cite{Vedantam_2015_CVPR, lin-2004-rouge, banerjee-lavie-2005-meteor, bleu_score}, several new AD metrics~\cite{Zhang_2024_CVPR, Han24, Han23} have been proposed, including CRITIC~\cite{Han24} measuring {\em who} is mentioned in the AD. 
However, these metrics fail to emphasise character {\em actions} -- one of the most critical aspects of ADs. To address this, we introduce an \textit{action score}, assessing whether the predictions accurately capture the actions described in GT AD, independent of character information.

Moreover, as ADs are time-limited, there is always a choice on which visual aspect (characters, actions, objects, etc.) should be included.
Consequently, a single video clip can have multiple equally valid ADs, each highlighting a different aspect. 
This observation is supported by the inter-rater agreement experiments in
AutoAD-III~\cite{Han24} and previous user studies~\cite{wang2021toward}. Therefore, beyond evaluating single AD performance, we assess our framework's capability as an \textit{assistant} to generate multiple AD candidates, and report the performance of the
selected \textit{one} AD.

In summary, we propose an enhanced training-free AD generation framework, with the following contributions:
\vspace{0.1cm}
\begin{itemize}[itemsep=0.2em]
    \item We incorporate shot-based temporal context into AD generation via training-free prompting techniques including shot number referral and dynamic frame sampling.
    \item We develop state-of-the-art methods for thread structure and shot scale predictions, and demonstrate that incorporating predicted film grammar knowledge enhances AD generation.
    \item We improve the current AD evaluation by introducing the character-free action score, and a new assistance-oriented evaluation protocol.
    \item Our approach achieves state-of-the-art performance in training-free AD generation, and furthermore, surpasses fine-tuned models on multiple benchmarks. This is the first time a training-free approach has achieved superior performance to fine-tuned methods.
\end{itemize}
\vspace{0.1cm}

\section{Related work}
\label{sec:rel_work}

\noindent \textbf{Audio Description generation. }
Efforts have been made to curate Audio Description (AD) datasets for both movies~\cite{soldan2022mad, Han24} and TV series~\cite{Xie24b}, with human annotations sourced from platforms such as AudioVault~\cite{audiovault}.  

For automatic AD generation, prior works~\cite{Han23, Han23a, Han24, movieseq, uniad, distinctad} fine-tune pre-trained VLMs on AD annotations to produce descriptions in an end-to-end manner. However, these methods face challenges due to limited high-quality AD annotations and the high computational cost of fine-tuning each new backbone. Alternatively, training-free approaches~\cite{Zhang_2024_CVPR, ye-etal-2024-mmad-multi, chu2024llmadlargelanguagemodel, Xie24b} have gained traction for their scalability and flexibility, allowing customised AD output based on official guidelines~\cite{ad_guideline} or specific needs. Yet, these methods still lag in performance, while our approach is the first to achieve results on par with fine-tuned methods.

Instead of limiting the video input to each AD interval, UniAD~\cite{uniad} and DistinctAD~\cite{distinctad} fine-tune VLMs with multiple AD clip inputs to incorporate broader temporal context. In contrast, our method systematically extends AD clip to adjacent shots and introduces a training-free approach that enables pre-trained VideoLLMs to better capture localised temporal context.

\vspace{2pt} \noindent \textbf{Film grammar analysis.}
Prior research has sought to understand film grammar from two major perspectives: (i)~intra-shot properties, (ii) shot-wise relationships.

For individual shots, several datasets~\cite{rao2020unified, argaw2022anatomy, MovieNet, TMM.2021.3092143, 3604321.3604334} categorise their characteristics based on camera setups, including shot scales (examples shown in~\cref{fig:shot_scale}) and camera movements. Correspondingly, various models~\cite{rao2020unified, Chen2022, Lu2024, electronics12194174, electronics11101570} have been developed for shot type classification.

Regarding shot-wise relationships, a few datasets~\cite{argaw2022anatomy, Chen_2023_WACV, pardo2022moviecuts} have been proposed to explore transitions (i.e.\ cuts) between shots, which have also been leveraged in video content generation~\cite{rao2022temporal, Zhang2023, 10447306, Pardo2024}. Beyond pairwise shot transitions, research has also investigated longer temporal contexts with thread-based editing structures. Notably, Hoai et al.\ introduced the Thread-Safe~\cite{Hoai14e} dataset, demonstrating that thread information can enhance action recognition. These structures have also been utilised in video-based face and people clustering~\cite{total_cluster, Brown21c}. 

While prior work on film grammar has mainly focused on classification and generation tasks, we specifically utilise shot scale and thread structure information to enhance AD generation in movies and TV series.

\vspace{2pt}
\noindent \textbf{Dense video captioning. }
Dense video captioning is closely related to AD generation. Early works~\cite{krishna2017dense,iashin2020better,iashin2020multi,wang2018bidirectional,wang2020event} in video captioning typically treat event localisation and captioning as independent stages, whereas more recent approaches~\cite{yang2023vid2seq,Perrett_2024_ACCV,Wu_2024_CVPR,Islam_2024_CVPR,Kim_2024_CVPR,Zhou_2024_CVPR,zhou2018end,wang2021end,chen2021towards,deng2021sketch,mun2019streamlined,rahman2019watch} integrate these tasks in an end-to-end manner.  Video captioning benchmarks cover a range of domains, including cooking~\cite{youcook2}, actions~\cite{krishna2017dense}, movies~\cite{rohrbach2015lsmdc}, TV series~\cite{lei2020tvr}, and open-domain settings~\cite{shvetsova2023howtocaption,miech2019howto100m,zhao2023lavila,zellers2022merlotreserve,xue2022hdvila,hendricks18emnlp,chen2024panda70m,chen2011msvd,xu2016msrvtt}.

\vspace{2pt} \noindent \textbf{Temporal grounding in VLM.}  
To equip conventional VLMs with temporal grounding capability, some studies~\cite{Huang_2024_CVPR,wang2024hawkeye,li2024groundinggpt} generate additional data with enhanced temporal information for fine-tuning. A more common approach explicitly incorporates temporal information into inputs, either by inserting temporal tokens~\cite{huang2024lita,hong2024cogvlm2,guo2024vtg,guo2024trace,qian2024momentor,wang2024grounded,chen2024timemarker,deng2024seq2time} or embedding time information within visual tokens~\cite{Huang_2024_CVPR,wang2024hawkeye,li2024groundinggpt}.  

Training-free approaches~\cite{Qu_2024_CVPR,zheng2024trainingfree,wu2024numberit} have also been explored for temporally grounded understanding. Notably, a recent work~\cite{wu2024numberit} achieves temporal grounding by overlaying frame numbers as visual prompts. Instead of using uniformly sampled timestamps, we leverage the natural shot structures in movies and TV series, adopting them as fundamental units for training-free temporal referral.

\vspace{2pt} \noindent \textbf{Evaluation for Audio Description. }
Early AD evaluation often adopts captioning metrics, including CIDEr~\cite{Vedantam_2015_CVPR}, ROUGE~\cite{lin-2004-rouge}, BLEU~\cite{bleu_score}, and METEOR~\cite{banerjee-lavie-2005-meteor}, which measure word/n-gram overlap with different weighting strategies. Other metrics focus on semantics, leveraging rule-based scene graph matching (SPICE~\cite{spice2016}), word embedding similarity (BERTScore~\cite{bert-score}, Recall@k/N~\cite{Han23}), or LLM-based evaluation~\cite{Han24, Zhang_2024_CVPR}.  

However, the fixed and distinct set of character names in each video can bias conventional captioning metrics. For instance, TF-IDF~\cite{Robertson2004} weighting in CIDEr assigns high importance to character names. To provide a more independent measure of character names and other AD content, CRITIC~\cite{Han24} assesses character recognition in predicted ADs, while we develop an action-centric metric with minimised dependence on character names. 

Most prior works evaluate single AD outputs; however, training-free methods are compatible with generating multiple candidates for selection. While early studies~\cite{wang2021toward} explored this through user studies, we propose a quantitative evaluation protocol to further assess AD generation methods as assistive tools.

\section{Training-free AD generation framework}
\label{sec:framework}
Given a video clip \(\mathcal{V} = \{\mathcal{I}_{0}, ..., \mathcal{I}_{T}\}\), the task of audio description is to generate a concise narration \(\mathcal{N}\) describing what happens around a given AD interval \([t_A, t_B]\). 
In this work, we propose a two-stage framework that leverages VideoLLMs and LLMs to predict ADs in a training-free manner. 

In Stage I, we employ a VideoLLM that takes a sequence of frames (from multiple shots) as input and generates a dense description \(\mathcal{D}\), guided by instructions \(\mathcal{P}_{\text{VideoLLM}}\):
\begingroup
\setlength{\abovedisplayskip}{7pt}
\setlength{\belowdisplayskip}{6pt}
\begin{align}
\mathcal{D} = \text{VideoLLM}(\mathcal{V}, [t_A, t_B], \mathcal{P}_{\text{VideoLLM}})
\end{align}
\endgroup

In Stage II, we then prompt an LLM (with instructions $\mathcal{P}_{\text{LLM}}$) to extract key information from the dense Stage I description and format it into an AD-style narration \(\mathcal{N}\):
\begingroup
\setlength{\abovedisplayskip}{8pt}
\setlength{\belowdisplayskip}{7pt}
\begin{align}
\mathcal{N} = \text{LLM}(\mathcal{D}, [t_A, t_B], \mathcal{P}_{\text{LLM}})
\end{align}
\endgroup

In this section, we focus on enhancing the visual understanding of the Stage I VideoLLM for edited video material, and make
three innovations: 
In~\cref{subsec:visual_context}, we incorporate shot-based temporal context into Stage I visual inputs. 
In~\cref{subsec:thread}, we leverage the thread structure to enrich cross-shot understanding. Finally, in~\cref{subsec:shot}, we incorporate shot-scale awareness into Stage I prompt formuation.

\subsection{Leveraging shot-based temporal context} 
Regarding the visual inputs to VideoLLM, prior works~\cite{movieseq,Han24} often 
sample frames \(\{\mathcal{I}_{t_A}, ..., \mathcal{I}_{t_B}\}\) that directly correspond to the AD interval \([t_A, t_B]\). However, this approach can be problematic due to (i) misalignment between the AD interval and the actual timestamps when the action occurs, and (ii) the lack of contextual information from adjacent shots. Therefore, we investigate how incorporating temporal context can enhance the understanding of the video clip.

\label{subsec:visual_context}
\begin{figure} 
\centering
\includegraphics[width=0.475\textwidth]{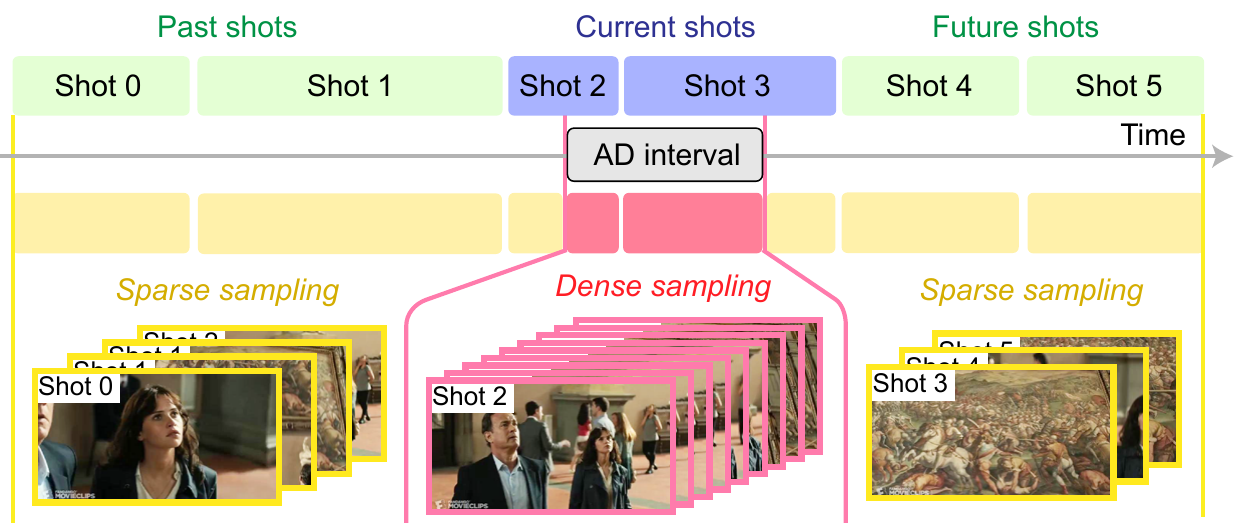}
\vspace{-0.65cm}
\caption{\small
\textbf{Shot-based temporal context,} where \textit{current shots} are defined as those temporally overlapping with the AD interval. \textit{Past shots} and \textit{future shots} provide extended contextual information. Shot numbers are visually overlaid on the top-left of each frame, and frames within the AD interval are sampled more densely than context frames.  
}
\vspace{-0.3cm}
\label{fig:visual_context}
\end{figure}

\vspace{2pt} \noindent \textbf{Structuring temporal context with shots.} 
To obtain more visual context information, instead of simply extending the AD interval by fixed timestamps, we explore a more structured approach that treats shots as the fundamental units.  

Specifically, we apply an off-the-shelf shot segmentation model to partition the entire video clip into individual shots. For each AD interval, we first identify the shots that (partially) overlap with it, referred to as ``current shots'', as illustrated in~\cref{fig:visual_context}.  We then consider \textit{at most} two ``past shots'' and two ``future shots'' adjacent to current shots as temporal context. For all shots included, we label them sequentially from past to future with a number starting from \textit{``Shot 0''}.

\vspace{2pt} \noindent \textbf{Emphasising the targeted AD interval.} 
To ensure that the VideoLLM focuses on describing visual content within the AD interval, we propose two strategies: dynamic frame sampling and shot number referral.

\vspace{2pt} \noindent \textit{Dynamic frame sampling.} 
As shown in~\cref{fig:visual_context}, to emphasise the frames of interest, we adopt denser sampling within the AD interval (red region) and sparser sampling for the surrounding context frames (yellow region). In practice, we specify fixed numbers of frames to be sampled within and outside the AD interval and apply uniform sampling according to these constraints.

\begin{figure}[t!]
\centering
\includegraphics[width=0.475\textwidth]{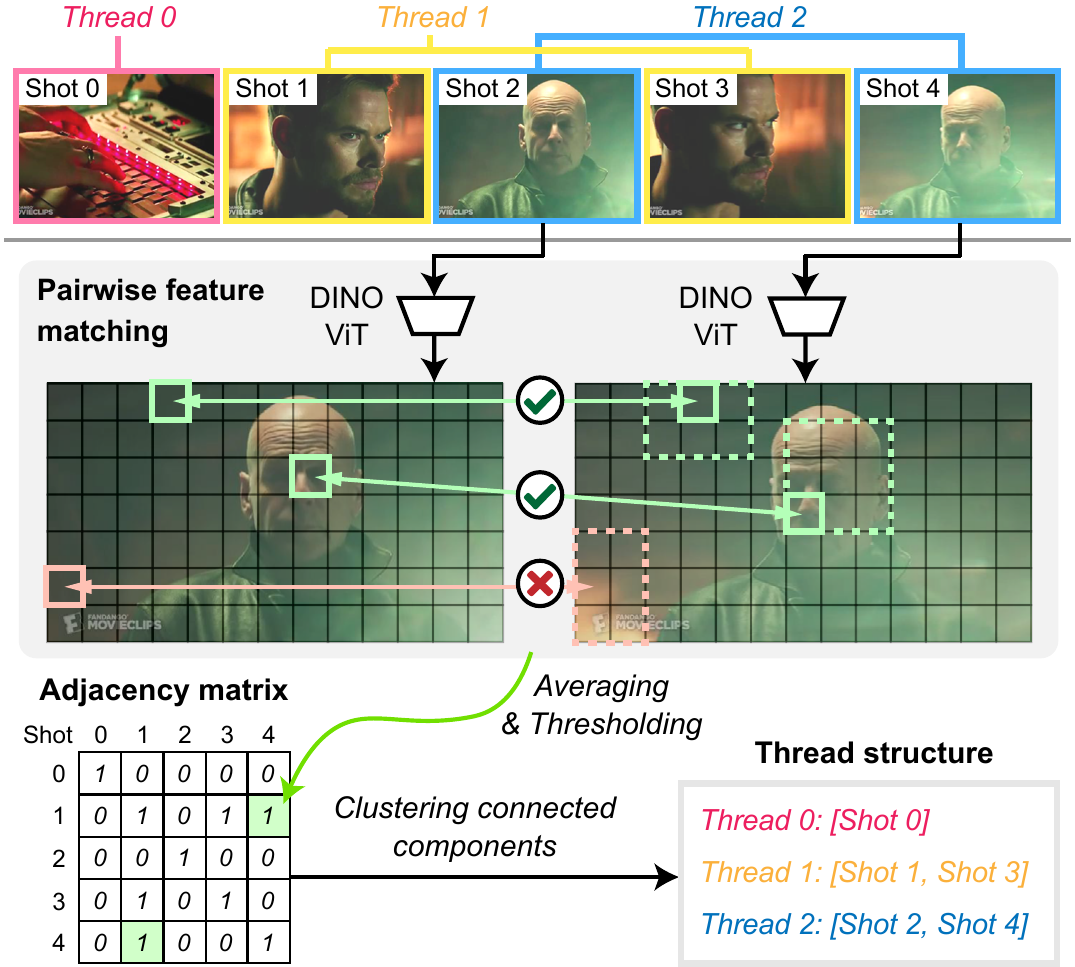}
\vspace{-0.6cm}
\caption{\small
\textbf{Thread structure.} \textbf{Top:} Example of a thread structure with interleaving shots. \textbf{Bottom:} A training-free approach for thread clustering, where shots are pairwise compared using dense feature matching to construct an adjacency matrix, which is then used to predict the thread structure.}
\label{fig:thread}
\vspace{-0.45cm}
\end{figure}

\vspace{2pt} \noindent \textit{Shot number referral.} 
To further enhance the attention towards the current shot content, we label each sampled frame with its shot number (e.g.\ \textit{``Shot 0''}) at the top-left. During the formulation of the text prompt, instead of prompting the VideoLLM to \textit{``describe what happened in the video clip''}, we ask it to \textit{``describe what happened in [Shot 2, Shot 3]''} (i.e.\ current shots). Through this visual-textual prompting strategy, we found that the VideoLLM could successfully interpret the meaning of shot numbers and refer to the correct shots.

\subsection{Leveraging thread structure}
\label{subsec:thread}

Movies are generally edited such that viewpoints from two or more cameras are intertwined in shot {\em threads}, as illustrated in ~\cref{fig:thread} (top). 
These interleaved arrangements of  shot threads often imply relationships between objects and characters (e.g.\ gaze interactions) and their 3D arrangement.
To leverage this information for Stage I description, we first determine the thread structure using a separate module, then incorporate it into the VideoLLM through prompt guidance.

\begin{figure}[t!]
\centering
\includegraphics[width=0.472\textwidth]{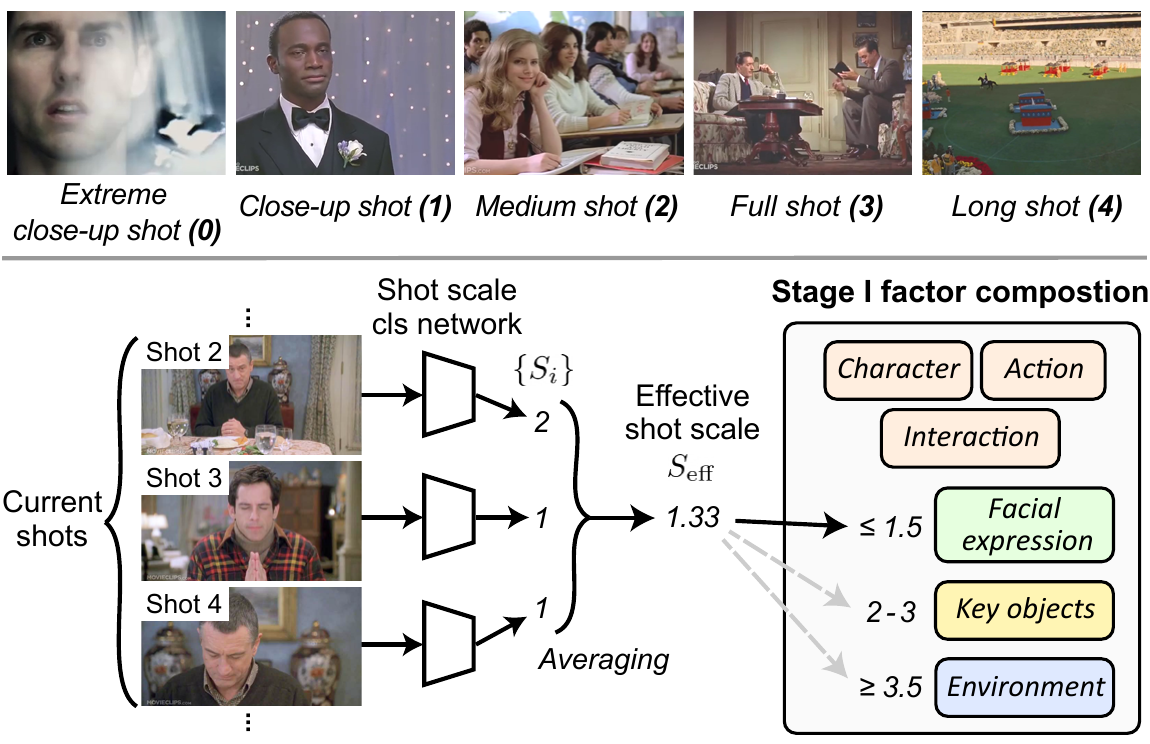}
\vspace{-0.2cm}
\caption{\small
\textbf{Shot scales.} \textbf{Top:} Examples of five different shot scales. \textbf{Bottom:} Stage I factor composition based on shot scale classification. The scales of the current shots in the clip are predicted and averaged. The resulting effective shot scale then guides the formulation of the Stage I prompt, incorporating additional factors such as facial expressions, etc.}    
\label{fig:shot_scale}
\vspace{-0.2cm}
\end{figure}

\begin{figure*}[t]
\centering
\includegraphics[width=0.925\textwidth]{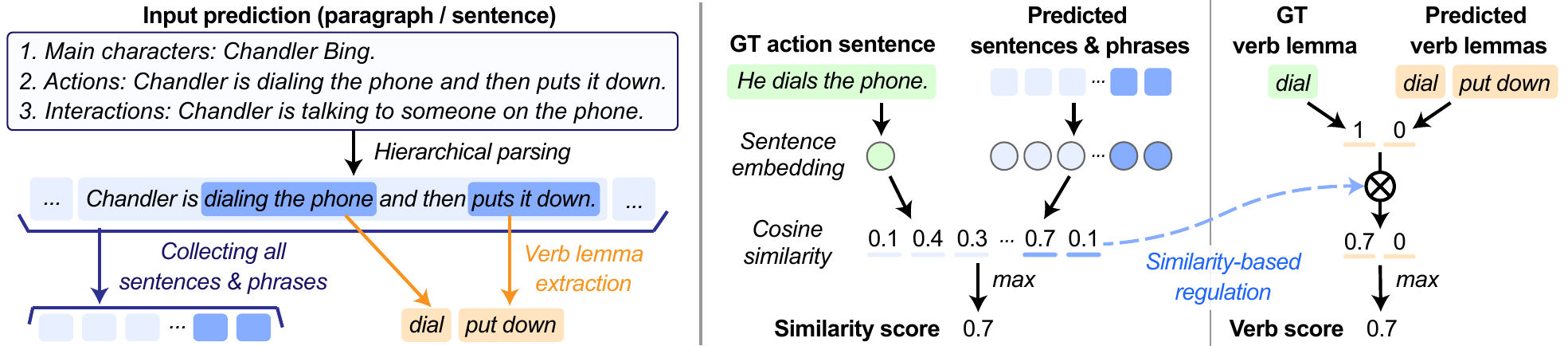}
\vspace{-0.2cm}
\caption{\small
\textbf{Action score.} \textbf{Left:} Hierarchical prediction parsing. Given a predicted paragraph or sentence, it is first divided into a set of sentences (light blue) and action phrases (dark blue). Each action phrase is further processed to obtain its verb lemma (light orange). 
\textbf{Middle:} Scoring based on semantic similarity. Sentence embeddings for the GT action sentence and the set of predicted sentences and phrases are extracted, with the maximum cosine similarity defined as the similarity score. \textbf{Right:} Scoring based on verb matching. The predicted verb lemma is compared with the GT lemma and multiplied by the corresponding semantic-based similarity. The highest resultant score is defined as the verb score.}
\vspace{-0.25cm}
\label{fig:action_score}
\end{figure*}

\vspace{2pt} \noindent \textbf{Thread structure prediction.} We develop a training-free method to predict thread clustering. The problem here is to determine if two shots {correspond} to the same viewpoint or not. Specifically, given two shots (Shot $i$ and a later Shot $j$), we compare the last frame of Shot $i$ (\(\mathcal{I}^{i}_{T_i}\)) with the first frame of Shot $j$ (\(\mathcal{I}^{j}_{0}\)), extracting their DINOv2~\cite{oquab2024dinov} features as \( f^{i}_{T_i} \) and \( f^{j}_{0} \in \mathbb{R}^{h\times w\times c} \).

Inspired by~\cite{jabri2020walk}, we assess frame-wise dense feature correlations by computing a cost volume between their feature maps:
\begingroup
\setlength{\abovedisplayskip}{8pt}
\setlength{\belowdisplayskip}{7pt}
\begin{align}
\mathcal{C}^{i,j}_{p,q} = m(p,q) \circ (\hat{f}^{i}_{T_i; p} \cdot \hat{f}^{j}_{0; q})
\end{align}
\endgroup
where \(\hat{f}^{i}_{T_i; p}\) and \(\hat{f}^{j}_{0; q} \in \mathbb{R}^{c} \) are the normalised $p$-th and $q$-th spatial elements in the respective feature maps. The binary attention mask \( m(p, q) \) is set to $1$ only if the spatial position of the $q$-th element is within an \( n \times n \) neighbourhood of the $p$-th element.

We then apply a softmax operation along the last dimension ($q$) and find the maximum similarity for each $p$, followed by averaging over all $p$-th elements to obtain a matching score between Shot $i$ and Shot $j$:
\begingroup
\setlength{\abovedisplayskip}{8pt}
\setlength{\belowdisplayskip}{7pt}
\begin{align}
s^{i,j}  = \frac{1}{N}\sum_{p}^{N}\max_q\left(\frac{\exp(\mathcal{C}^{i,j}_{p,q}/ \tau)}{\sum_{l}^{N}\exp(\mathcal{C}^{i,j}_{p,l}/ \tau )}\right)
\end{align}
\endgroup
where \( N \) denotes the number of feature patches, and \( \tau \) is the softmax temperature.

Intuitively, as shown in~\cref{fig:thread}, this process effectively checks whether each patch ($p$) in one shot frame matches with a patch ($q$) in the other shot frame at a roughly similar spatial position (i.e., within an \( n\times n \) neighbourhood).  

Finally, we construct an adjacency matrix based on the predicted scores \( s^{i,j} \) for all possible pairs of Shot $i$ and Shot $j$. By thresholding the adjacency matrix (threshold $\epsilon$) and identifying the largest connected components, we cluster the set of shots into multiple threads.

\vspace{2pt} \noindent \textbf{Thread structure injection.} 
Once the thread information is obtained, we inject it into the text prompt \(\mathcal{P}_{\text{VideoLLM}}\) for the Stage I VideoLLM. In practice, we conduct this information injection only to clips that exhibit thread structures (i.e.\ \(N_{\text{thread}} < N_{\text{shot}}\)). For each given thread \([\text{Shot } i, ..., \text{Shot } j]\), we formulate the prompt as:  
\textit{``$[\text{Shot } i, ..., \text{Shot } j]$ share the same camera setup''.}
This statement implies that the cameras in these shots maintain consistent angles and scales.

Rather than simply providing this information, we further engage the VideoLLM by asking it to explain why the given thread structure is correct. This effectively corresponds to a Chain-of-Thought (CoT) process, enhancing its understanding of these repetitive thread structures.

\subsection{Leveraging shot scale information}
\label{subsec:shot}

In movies and TV series, shot scales are often carefully designed during filming or post-editing to implicitly convey information to the audience. 
Our objective is to use the  shot scale to choose what should be included in the Stage~I prompts.

\vspace{2pt} \noindent \textbf{Shot scale classification.}
We first build a shot scale classification network by fine-tuning a pre-trained DINOv2~\cite{oquab2024dinov} model.  
For all \textit{current} shots, we classify their shot scales \(\{S_i\}\) into one of \textit{five} classes, represented by values $0$–$4$, as shown in~\cref{fig:shot_scale} (top). We then compute their average to obtain the effective shot scale \(S_\text{eff}\).

\vspace{2pt} \noindent \textbf{Stage I factor composition.} 
We then leverage the predicted shot scale to determine the factors to include in Stage I instructions. We first consider three fixed factors that form the basis of ADs, namely characters, actions, and interactions. The additional factor can be determined through applying a set of thresholds to \(S_\text{eff}\), as detailed in~\cref{fig:shot_scale} (bottom). For example, for close-up shots (\(S_\text{eff} \leq 1.5\)), we would ask the VideoLLM to additionally describe the ``facial expression'', whereas for long shots (\(S_\text{eff} \geq 3.5\)), a description on the ``environment'' will be included.

\section{Action score}
\label{sec:action_score}
In this section, we introduce a new metric, termed ``action score'', which focuses on whether a specific ground truth (GT) action is captured within a predicted Stage I description paragraph or a Stage II AD output. 
For instance, for the GT action \textit{``He dials the phone''}, we want the metric to measure the action of \textit{``dial the phone''}, but not be sensitive to character names and other predicted content.
Therefore, this metric is designed to possess two key properties: (i) it is character-free, meaning that the presence of character names has minimal impact on the evaluation, and (ii) it is recall-oriented, without penalising additional action information in the prediction.

\vspace{2pt} \noindent \textbf{Preprocessing of GT actions.} 
For each GT AD, we extract the character-free GT actions by (i) replacing character names with pronouns, and (ii) splitting the AD into subsentences, each containing one action (verb). For example, given a GT AD \textit{``Chandler dials the phone, then hurriedly hangs up.''}, the extracted GT actions are \textit{``He dials the phone.''} and \textit{``He hurriedly hangs up.''}

\vspace{2pt} \noindent \textbf{Hierarchical parsing of predictions.} 
To process the predicted paragraph during evaluation, we first decompose it into individual sentences (light blue), as shown in the left column of~\cref{fig:action_score}. For each predicted sentence, we then perform rule-based dependency parsing to obtain action phrases (deep blue). Then, all extracted sentences and action phrases are collected to form a prediction set. Additionally, for each action phrase, we extract the corresponding verb lemma (i.e.\ the root form of verbs).

\vspace{2pt} \noindent \textbf{Action score computation.} 
The action score features a combination of semantic-based and verb-matching-based components, as illustrated in the middle and right columns of~\cref{fig:action_score}, respectively.

\vspace{1pt} \noindent \textit{Semantic-based evaluation.} 
For each GT action sentence, we assess its semantic similarity with the set of predicted sentences and phrases. Specifically, we employ a general text embedding (GTE) model to compute \textit{sentence-level} embeddings for the GT action sentence \( e_{\text{GT}} \) and each element in the predicted set \( \{e_{\text{pred}; i}\} \). By computing cosine similarities and taking the maximum value, {the \textit{similarity score} is defined as \( s_{\text{sim}} = \max_i(s_{\text{sim}; i}) = \max_i[({e_{\text{GT}}\cdot e_{\text{pred}; i}})/({\|e_{\text{GT}}\|\|e_{\text{pred}; i}\|})] \).}

\vspace{1pt} \noindent  \textit{Verb-matching-based evaluation.} 
In addition to semantic matching, we credit predictions that contain the same verbs as those in GT actions. Practically, we compare the predicted verb lemmas with GT verb lemma and compute binary matching scores \( \{m_j\} \). These matching scores are further weighted by the corresponding similarity scores \( \{s_{\text{sim}, j}\} \) through element-wise multiplication. 
Finally, we take the maximum of the regulated scores to compute the \textit{verb score}, i.e.\ \( s_{\text{verb}} = \max_{j}(s_{\text{sim};j} \circ m_j) \).

To obtain the final action score, we combine the scores from both sources using a weighted average \(s_{\text{action}} = \alpha_{\text{sim}} \circ s_{\text{sim}} + \alpha_{\text{verb}} \circ s_{\text{verb}} \), with \(\alpha_{\text{sim}}\) and \(\alpha_{\text{verb}}\) denoting the weighted factors.
\section{Experiments}
\label{sec:exp}
This section begins with datasets and metrics for AD generation in~\cref{subsec:ad_dataset_metric}, followed by implementation details in~\cref{subsec:imple}. Results on film grammar predictions are presented in~\cref{subsec:exp_thread,subsec:exp_scale}, while \cref{subsec:human_agree} analyses human alignment with action scores. AD generation results are detailed in~\cref{subsec:ad_components,subsec:ad_compare,subsec:ad_assist}.

\subsection{AD generation datasets and metrics}
\label{subsec:ad_dataset_metric}
We evaluate our framework on AD generation datasets for both movies (CMD-AD~\cite{Han24}, MAD-Eval~\cite{Han23}) and TV series (TV-AD~\cite{Xie24b}).
In more detail, CMD-AD is constructed by aligning ground truth ADs with the Condensed Movie Dataset (CMD)~\cite{Bain20}, comprising $101$k ADs ($94$k for training and $7$k for testing) from $1.4$k movies. MAD-Eval consists of $6.5$k ADs sampled from $10$ movies within LSMDC~\cite{rohrbach2015lsmdc}. On the other hand, TV-AD features $34$k AD annotations from $13$ TV series, with its test set sourced from TVQA~\cite{lei2018tvqa}, consisting of $3$k ADs.

For AD evaluation, we follow the prior works~\cite{Han24,distinctad} to assess general prediction quality using CIDEr~\cite{Vedantam_2015_CVPR}, Recall@k/N~\cite{Han23a}, and LLM-AD-Eval~\cite{Han24}. We also consider character recognition accuracy using CRITIC~\cite{Han24} and character-free action evaluation using action scores (described in~\cref{sec:action_score}).

\subsection{Implementation details}
\label{subsec:imple}
In this section, we provide key implementation details for AD generation and action score computation. For additional information on film grammar predictions and other details, please refer to~\cref{supsec:imple}.

\vspace{2pt} \noindent \textbf{Shot detection.}  
To segment the video clip into shots, we use PySceneDetect~\cite{scenedetect} with the ``Adaptive Detection'' method, 
which compares the ratio of pixel changes with the neighbouring frames. On average, each shot spans $3.5$s, $3.3$s, $3.8$s in CMD-AD, TV-AD, and MAD-Eval, respectively.

\vspace{2pt} \noindent \textbf{AD generation setting.}  
During dynamic frame sampling, we select a total of $32$ frames, with $16$ frames uniformly sampled within and outside the AD interval, respectively. For character recognition, we adopt the same visual-textual prompting method proposed by~\cite{Xie24b}, which applies coloured circles around faces for visual character indication. For simplicity, the visualisations in this paper do not display these circle labels.

Regarding the base models, we use Qwen2-VL-7B~\cite{Qwen2VL} as the VideoLLM in Stage I and LLaMA3-8B~\cite{grattafiori2024llama3herdmodels} as the LLM in Stage II. This setup is used as the \textit{default} unless stated otherwise. Additionally, we explore the framework with the proprietary GPT-4o~\cite{openai2024gpt4technicalreport} model for both stages.

\vspace{2pt} \noindent \textbf{Action score evaluation setting.}  
For action score computation, we set the weight factors as \( \alpha_{\text{sim}}\) = \( 0.8 \) and \( \alpha_{\text{verb}}\) = \( 0.2 \). When aggregating the action score results, we first average over multiple GT actions within each GT AD, and then perform global averaging across all AD samples. Moreover, in practice, we find that most action scores are clustered within the range of $0.25 - 0.75$. To improve clarity, we apply further rescaling \( f(x) = (x - 0.25) \times 2 \) as post-processing. For evaluations in this paper, unless otherwise specified, we use the action score to assess the Stage II AD outputs.

\begin{table}[t]
\centering
\setlength\tabcolsep{6.5pt}
\resizebox{0.48\textwidth}{!}{
\begin{tabular}{lccccc}
\toprule
~~~~~~~Feature & Frame setup & Precision & Recall & AP & WCP \\ 
\midrule
CLIP-L14 CLS & Side &  $0.691$ & $0.635$ & $0.705$ & $0.922$ \\ 
DINOv2-L14 CLS & Side  & $0.759$ & $0.683$ & $0.788$ & $0.933$ \\ 
DINOv2-g14 CLS & Side  & $0.761$ & $0.675$ & $0.786$ & $0.936$ \\ 
\midrule
DINOv2-g14 spatial& Mid & $0.808$ & $0.717$ & $0.822$ & $0.953$ \\ 
DINOv2-g14 spatial& All & $0.870$ & $0.795$ & $0.896$ & $0.964$ \\ 
\midrule
DINOv2-g14 spatial& Side  & $\mathbf{0.878}$ & $\mathbf{0.799}$ & $\mathbf{0.902}$ & $\mathbf{0.965}$ \\ 
\bottomrule
\end{tabular}
}
\vspace{-0.25cm}
\caption{\small\textbf{Thread structure prediction on Thread-Safe.} ``Side'' refers to comparisons between the temporally nearest frames in two shots; ``Mid'' refers to comparisons between the middle frames of each shot; ``All'' refers to comparisons between all frame pairs from the two shots.}
\vspace{-0.35cm}
\label{tab:thread_structure}
\end{table}

\subsection{Thread structure prediction}
\label{subsec:exp_thread}
We evaluate thread structure prediction on Thread-Safe~\cite{Hoai14e}, which consists of approximately $4.7$k video clips collected from $15$ TV series. Each video clip contains a multi-shot scene with corresponding thread clusters manually annotated.

For evaluation, we first construct an adjacency matrix from the GT clusters and extract binary GT labels \( \hat{s}_{i,j} \) from the off-diagonal entries, where each label indicates whether a given pair of shots belong to the same cluster. We then compute the Average Precision (AP) between the predicted relationships \( \{s_{i,j}\} \) and the ground truth \( \{\hat{s}_{i,j}\} \), as well as report the precision and recall values. Additionally, we directly compare the predicted clusters with ground truth clusters by reporting the weighted clustering purity (WCP)~\cite{total_cluster}.

\cref{tab:thread_structure} verifies the choice of DINOv2-g14~\cite{oquab2024dinov} as the feature extractor for frame pair comparison. Compared to abstract CLS tokens, dense spatial matching achieves higher AP in frame pair relationship prediction and higher WCP in thread clustering. Additionally, we observe that using the temporally closest frames from two shots (``Side'') leads to a noticeable performance improvement. This can be attributed to the continuous story flow across shots within the same thread.

\subsection{Shot scale classification}  
\label{subsec:exp_scale}
Following prior work on shot scale classification~\cite{Lu2024, electronics12194174}, we use the MovieShots~\cite{rao2020unified} dataset, which consists of $46$k shots (train:val:test = $7$:$1$:$2$) collected from over $7$k movie trailers. We follow its definition of shot scales, categorising shots into five classes ranging from extreme close-up to long shots, as illustrated in~\cref{fig:shot_scale} (top). To evaluate the model performance, we report classification accuracy and Macro-F1~\cite{electronics12194174} scores on the MovieShots test set.

Since previous state-of-the-art methods on shot scale classification are not open-sourced, we develop a new network by fine-tuning DINOv2~\cite{oquab2024dinov}, achieving superior performance over prior approaches that rely on additional optical flow or SAM-based mask inputs, as demonstrated in~\cref{tab:shot_scale}.

\begin{table}[t]
\centering
\setlength\tabcolsep{10pt}
\resizebox{0.4\textwidth}{!}{
\begin{tabular}{lccc}
\toprule
Metric & Input & Accuracy & Macro-F1 \\ 
\midrule
ViViT~\cite{Arnab_2021_ICCV} & RGB & $0.747$ & $0.751$ \\ 
SGNet~\cite{rao2020unified} & RGB + Flow & $0.875$ & $-$ \\ 
Lu et al.~\cite{Lu2024} & RGB + mask & $0.892$ & $-$ \\ 
Li et al.~\cite{electronics12194174} & RGB & $0.895$ & $0.897$ \\ 
\midrule
\textbf{Ours} & RGB & $\mathbf{0.897}$ & $\mathbf{0.899}$ \\ 
\bottomrule
\end{tabular}
}
\vspace{-0.25cm}
\caption{\small\textbf{Shot scale classification on MovieShots.}}
\label{tab:shot_scale}
\vspace{-0.2cm}
\end{table}

\begin{table}[t]
\centering
\setlength\tabcolsep{5pt}
\resizebox{0.48\textwidth}{!}{
\begin{tabular}{lcccc}
\toprule
\multirow{2}{*}{Metric} & \multicolumn{2}{c}{Paragraph}  &  \multicolumn{2}{c}{Sentence} \\ 
\cmidrule(lr){2-3} \cmidrule(lr){4-5}
 & Pearson & Spearman & Pearson & Spearman \\
\midrule
CIDEr~\cite{Vedantam_2015_CVPR} & $0.205$ & $0.264$ & $0.412$ & $0.528$\\ 
ROUGE-L~\cite{lin-2004-rouge} & $0.305$ & $0.280$ & $0.526$ & $0.512$ \\ 
METEOR~\cite{banerjee-lavie-2005-meteor} & $0.462$ & $0.406$ & $0.602$ & $0.641$ \\ 
BLEU-1~\cite{bleu_score} & $0.265$ & $0.264$ & $0.477$ & $0.481$ \\ 
SPICE~\cite{spice2016} & $0.022$ & $0.048$ & $0.031$ & $0.012$ \\ 
BERTScore~\cite{bert-score} & $0.377$ & $0.393$ & $0.508$ & $0.507$ \\ 
LLM-based (LLaMA3.1-70B~\cite{grattafiori2024llama3herdmodels}) & $0.569$ & $0.491$ & $0.779$ & $0.798$\\ 
LLM-based (GPT-4o~\cite{openai2024gpt4technicalreport}) & $0.742$ & $0.678$ & $0.797$ & $0.807$\\ 
\midrule
\textbf{Action Score} (w/o verb matching) & $0.735$  & $0.728$  & $0.765$ & $0.790$ \\ 
\textbf{Action Score} (w verb matching) & $\mathbf{0.749}$ & $\mathbf{0.729}$ & $\mathbf{0.806}$ & $\mathbf{0.820}$ \\ 
\bottomrule
\end{tabular}
}
\vspace{-0.25cm}
\caption{\small\textbf{Comparison of action score with other metrics.} The listed metrics measure the similarity between predicted paragraphs/sentences and ground truth actions. The reported values indicate the correlation (i.e.\ alignment) between these metrics and human-annotated scores.
}
\label{tab:action_scores}
\vspace{-0.35cm}
\end{table}

\subsection{Human alignment with action scores} 
\label{subsec:human_agree}
Action scores aim to evaluate whether a GT action is captured within a predicted description, making them recall-oriented. Such descriptions can be in the form of paragraphs (Stage I descriptions) or single sentences (Stage II ADs).
{To assess whether action scores align with human judgments, we create a dataset containing pairs of predicted descriptions and GT actions. For each GT action, human annotators \emph{manually annotate} the quality of predictions into $\{0,1,2,3\}$ based on the relevance towards GT action ranging from ``unrelated'' (0) to ``exact matching'' (3).}

\vspace{2pt} \noindent \textbf{Comparison with other metrics.}
Next, we use the human-annotated scores as a reference to compare different metrics in terms of human agreement (measured by correlations), as reported in~\cref{tab:action_scores}.
Additionally, we consider an LLM-based metric to predict scores, following the same scoring criteria as human annotations. 
In general, the action score achieves the best correlations with human annotations. Note, it is also more efficient than LLM-based metrics, with $0.15$s compared to $6$s per prediction evaluation. For more details regarding this human agreement study, please refer to~\cref{supsec:action_score}.

\begin{table}[t]
\centering
\setlength\tabcolsep{2pt}
\resizebox{0.48\textwidth}{!}{
\begin{tabular}{cccccccccc}
\toprule
\multirow{2}[2]{*}{Exp.} & \multirow{2}[2]{*}{\shortstack{Temporal \\ context}} & \multirow{2}[2]{*}{\shortstack{Frame \\ sampling}} & \multirow{2}[2]{*}{Shot label} & \multicolumn{3}{c}{CMD-AD} & \multicolumn{3}{c}{TV-AD} \\ 
\cmidrule(lr){5-7} \cmidrule(lr){8-10}
& & & &  CIDEr & CRITIC & Action & CIDEr & CRITIC & Action \\ 
\midrule
A & \cellcolor{lightblue}$-$ & $-$ & $-$ & $22.4$ & $45.7$ & $27.0$ & $26.0$ & $42.2$ & $22.3$ \\ 
B & \cellcolor{lightblue}$1$ shot & Dyn. & Shot num. & $24.7$ & $46.8$ & $\mathbf{27.8}$ & $\mathbf{28.9}$ & $41.6$ & $\mathbf{23.1}$ \\ 
\midrule
C & $2$ shots & Dyn. & \cellcolor{lightblue}$-$ & $24.5$ & $46.6$ & $27.4$ & $27.8$ & $\mathbf{42.3}$  & $22.3$\\ 
D & $2$ shots & Dyn. & \cellcolor{lightblue}\;Frame box\; & $24.8$ & $46.4$ & $27.4$ & $25.0$ & $41.8$ & $22.8$ \\ 
\midrule
E & $2$ shots & \cellcolor{lightblue}Uni. & Shot num. & $24.1$ & $47.5$ & $26.2$ & $26.5$ & $42.2$  & $22.0$\\ 
\textbf{F} & $2$ shots & Dyn. & Shot num. & $\mathbf{25.1}$ & $\mathbf{47.5}$ & $\mathbf{27.8}$ & $\mathbf{28.9}$ & $42.1$ & $23.0$  \\ 
\bottomrule
\end{tabular}
}
\vspace{-0.25cm}
\caption{\small \textbf{Leveraging shot-based temporal context.} Key changes relative to the default setting (Exp.\ F) are highlighted in light blue. ``Temporal context'' indicates the number of context (past \& future) shots. ``Shot num.'' refers to overlaying the shot number at the top-left of each current shot frame, while ``frame box'' represents highlighting the boundary of each current shot frame with a red box.}
\label{tab:temporal_context}
\vspace{-0.15cm}
\end{table}

\begin{table}[t]
\centering
\setlength\tabcolsep{8pt}
\resizebox{0.47\textwidth}{!}{
\begin{tabular}{ccccccccc}
\toprule
\multirow{2}[2]{*}{\shortstack{Thread\\structure}} & \multicolumn{3}{c}{CMD-AD subset} & \multicolumn{3}{c}{TV-AD subset} \\ 
\cmidrule(lr){2-4} \cmidrule(lr){5-7}
 & CIDEr & CRITIC & Action & CIDEr & CRITIC & Action \\ 
\midrule
\addlinespace[0.35em]
\xmark & $29.9$ & $47.5$ & $27.4$  & $28.8$ & $42.0$ & $22.6$ \\ 
\addlinespace[0.15em]
$\checkmark$ &  $\mathbf{30.7}\garrow{0.8}$ & $\mathbf{48.9}\garrow{1.4}$ & $\mathbf{27.7}\garrow{0.3}$  & $\mathbf{30.7}\garrow{1.9}$ & $\mathbf{42.7}\garrow{0.7}$ & $\mathbf{22.9}\garrow{0.3}$  \\ 
\addlinespace[0.05em]
\bottomrule
\end{tabular}
}
\vspace{-0.25cm}
\caption{\small\textbf{Thread structure injection.} Thread structure information is injected only into subsets predicted to exhibit thread structures ($\sim\!30\%$ in CMD-AD and $\sim\!60\%$ in TV-AD).}
\label{tab:thread_structure_injection}
\vspace{-0.15cm}
\end{table}

\begin{table}[t]
\centering
\setlength\tabcolsep{3pt}
\resizebox{0.48\textwidth}{!}{
\begin{tabular}{lcccccccc}
\toprule
\multirow{2}[2]{*}{Stage I factors} & \multicolumn{3}{c}{CMD-AD} & \multicolumn{3}{c}{TV-AD} \\ 
\cmidrule(lr){2-4} \cmidrule(lr){5-7}
 & CIDEr & CRITIC & Action &  CIDEr & CRITIC & Action \\ 
\midrule
Base & $25.4$ & $47.4$ & $27.7$ & $29.2$ & $42.0$ & $23.0$ \\ 
Base + Face (AutoAD-Zero) & $25.2$ & $46.8$ & $27.8$ & $30.0$ & $42.3$ & $22.5$ \\ 
Base + Obj. & $26.1$ & $45.8$ & $27.9$ & $30.4$ & $\mathbf{42.9}$ & $22.8$ \\ 
Base + Env. & $25.2$ & $46.7$ & $27.4$ & $29.8$ & $40.7$ & $22.2$ \\ 
Base + Face + Obj. + Env. & $26.0$ & $47.2$ & $27.4$ & $30.1$ & $42.7$ & $22.9$\\ 
\midrule
\textbf{Scale-dependent (Ours)} & $\mathbf{26.3}$ & $\mathbf{47.8}$ & $\mathbf{28.4}$ & $\mathbf{31.1}$ & $42.2$ & $\mathbf{23.9}$\\ 
\bottomrule
\end{tabular}
}
\vspace{-0.25cm}
\caption{\small\textbf{Factors included in Stage I description.}
Base: character + action + interaction; Face: facial expression; Env.: environment; Obj.: object. ``Scale-dependent'' refers to our approach, which leverages shot scale predictions to determine the relevant factors for each clip.}
\label{tab:scale_dependent}
\vspace{-0.3cm}
\end{table}

\begin{table*}[hpbt!]
\centering
\setlength\tabcolsep{3pt}
\resizebox{0.995\textwidth}{!}{
\begin{tabular}{lcccccccccccccc}  
\toprule
\multirow{2}[2]{*}{Method} &\multirow{2}[2]{*}{VLM} & \multirow{2}[2]{*}{LLM} & \multirow{2}[2]{*}{\shortstack{Training \\ -free}}  & \multirow{2}[2]{*}{\shortstack{Propriet. \\ model}} &  \multicolumn{5}{c}{CMD-AD} & \multicolumn{5}{c}{TV-AD}  \\
\cmidrule(r){6-10}
\cmidrule(r){11-15}
 &  & & & & CIDEr & CRITIC & Action & R@1/5  & LLM-AD-Eval & CIDEr & CRITIC & Action & R@1/5 & LLM-AD-Eval \\
\midrule
AutoAD-II~\cite{Han23a} & CLIP-B32 & GPT-2 & \xmark  & \xmark  & $13.5$ & $8.2$ & $-$ & $26.1$ & $2.08  \;|\;\;\;\;\,\; - \;\;\;\;$ & $-$ & $-$ & $-$ & $-$ & $-$ \\
AutoAD-III~\cite{Han24} & EVA-CLIP & LLaMA2-7B & \xmark  & \xmark & $25.0$ & $32.7$ & $31.5$ & $31.2$ & $2.89 \;|\; 2.01$ & $26.1$ & $28.8$ & $26.4$ & $30.1$ & $2.78 \;|\; 1.99$ \\
DistinctAD~\cite{distinctad} & CLIP$_{\text{AD}}$-B16 & LLaMA3-8B & \xmark  & \xmark & $22.7$ & $-$ & $-$ & $33.0$ & $2.88 \;|\;  2.03$ & $27.4$ & $-$ & $-$ & $32.1$ & $2.89 \;|\; 2.00$ \\
\midrule
Video-LLaMA~\cite{damonlpsg2023videollama} & Video-LLaMA-7B & $-$ & $\checkmark$ & \xmark & $4.8$ & $0.0$ & $-$ & $22.0$ & $1.89  \;|\;\;\,\;\;\; - \;\;\;\;$ & $-$ & $-$ & $-$ & $-$ & $-$  \\
VideoBLIP~\cite{videoblip} & VideoBLIP & $-$ & $\checkmark$ &  \xmark & $5.2$ & $0.0$ & $-$ & $23.6$ & $1.91 \;|\;\;\;\;\,\; - \;\;\;\;$ & $-$ & $-$ & $-$ & $-$ & $-$ \\
AutoAD-Zero~\cite{Xie24b} & VideoLLaMA2-7B & LLaMA3-8B & $\checkmark$ & \xmark & $17.7$ & $43.7$ & $25.5$ & $26.9$ & $2.83 \;|\; 1.96$ & $22.6$ & $39.4$ & $21.7$ & $27.4$ & $2.94 \;|\; 2.00$ \\
AutoAD-Zero~\cite{Xie24b} & Qwen2-VL-7B & LLaMA3-8B & $\checkmark$ & \xmark & $21.9$ & $44.3$ & $26.9$ & $30.8$ & $3.00 \;|\; 2.20$ & $26.4$ & $41.6$ & $22.1$ & $30.4$ & $3.05  \;|\;  2.27$ \\
\textbf{Ours} &  Qwen2-VL-7B & LLaMA3-8B  & $\checkmark$ & \xmark & $\mathbf{26.3}$ & $47.8$ & $28.4$ & $33.0$ & $3.15 \;|\; 2.42$ & $31.1$ & $42.2$ & $23.9$ & $33.1$ & $3.09  \;|\;  2.35$ \\
\midrule
AutoAD-Zero~\cite{Xie24b} & GPT-4o & GPT-4o & $\checkmark$ & $\checkmark$  & $22.4$ & $45.1$ & $30.7$ & $32.9$ & $3.08 \;|\;  2.49$ & $30.9$ & $44.4$ & $26.8$ & $34.7$ & $\mathbf{3.12} \;|\; 2.57$ \\
\textbf{Ours} & GPT-4o & GPT-4o & $\checkmark$ & $\checkmark$  & $26.1$ & $\mathbf{49.1}$ & $\mathbf{32.5}$ & $\mathbf{36.5}$ & $\mathbf{3.17} \;|\; \mathbf{2.66}$ & $\mathbf{34.2}$ & $\mathbf{46.5}$ & $\mathbf{27.4}$ & $\mathbf{36.6}$ & $\mathbf{3.12} \;|\;  \mathbf{2.59}$ \\ 
\bottomrule
\end{tabular}}
\vspace{-0.2cm}
\caption{\small\textbf{Quantitative comparison on CMD-AD and TV-AD.} For training-free methods, ``VLM'' and ``LLM'' refer to the models used in separate stages, while for fine-tuned models, they denote the pre-trained components within an end-to-end model.} 
\label{tab:quantitative}
\end{table*}

\begin{table*}[t]
\centering
\setlength\tabcolsep{6pt}
\resizebox{0.94\textwidth}{!}{
\begin{tabular}{lccccccccccc}
\toprule
\multirow{2}[2]{*}{Method} & \multirow{2}[2]{*}{VLM} & \multirow{2}[2]{*}{LLM} & \multirow{2}[2]{*}{\shortstack{Training \\ -free}}  & \multirow{2}[2]{*}{\shortstack{Propriet. \\ model}} & \multicolumn{6}{c}{MAD-Eval} \\
\cmidrule(r){6-11}
 &  &  &  &  & CIDEr & R@5/16 & Rouge-L & SPICE & METEOR & BLEU-1 \\
\midrule
ClipCap~\cite{mokady2021clipcap} & CLIP-B32 & GPT-2 & \xmark & \xmark & $4.4$ & $36.5$ & $8.5$ & $1.1$ & $-$ & $-$ \\ 
CapDec~\cite{capdec} & $-$ & $-$ & \xmark & \xmark & $6.7$ & $-$ & $8.2$ & $1.4$ & $-$ & $-$ \\ 
AutoAD-I~\cite{Han23} & CLIP-B32 & GPT-2 & \xmark & \xmark & $13.4$ & $42.1$ & $11.9$ & $4.4$ & $-$ & $-$ \\ 
AutoAD-II~\cite{Han23a} & CLIP-B32 & GPT-2 & \xmark & \xmark & $19.5$ & $51.3$ & $13.4$ & $-$ & $-$ & $-$ \\ 
AutoAD-III~\cite{Han24} & EVA-CLIP & LLaMA2-7B & \xmark & \xmark & $24.0$ & $52.8$ & $13.9$ & $6.1$ & $5.5$ & $13.1$ \\ 
MovieSeq~\cite{movieseq} & CLIP-B16 & LLaMA2-7B & \xmark & \xmark & $24.4$ & $51.6$ & $15.5$ & $7.0$ & $-$ & $-$ \\ 
DistinctAD~\cite{distinctad} & CLIP$_{\text{AD}}$-B16 & LLaMA3-8B & \xmark & \xmark & $27.3$ & $56.0$ & $\mathbf{17.6}$ & $8.3$ & $-$ & $-$ \\ 
UniAD~\cite{uniad} & CLIP-L14 & LLaMA-8B & \xmark & \xmark & $\mathbf{28.2}$ & $54.9$ & $17.2$ & $-$ & $-$ & $-$ \\ 
\midrule
Video-LLaMA~\cite{damonlpsg2023videollama} & Video-LLaMA-7B & $-$ & $\checkmark$ & \xmark & $4.8$ & $33.8$ & $-$ & $-$ & $-$ & $-$ \\ 
Video-BLIP~\cite{videoblip} & Video-BLIP & $-$ & $\checkmark$ & \xmark & $5.0$ & $35.2$ & $-$ & $-$ & $-$ & $-$ \\ 
AutoAD-Zero~\cite{Xie24b} & VideoLLaMA2-7B & LLaMA3-8B & $\checkmark$ & \xmark & $22.4$ & $47.0$ & $14.4$ & $7.3$ & $6.6$ & $15.1$ \\ 
AutoAD-Zero~\cite{Xie24b} & Qwen2-VL-7B & LLaMA3-8B & $\checkmark$ & \xmark & $23.6$ & $51.3$ & $14.6$ & $7.8$ & $6.6$ & $13.6$ \\ 
\textbf{Ours} & Qwen2-VL-7B & LLaMA3-8B & $\checkmark$ & \xmark & $25.0$ & $50.6$ & $14.7$ & $7.8$ & $7.2$ & $\mathbf{16.2}$ \\ 
\midrule
VLog~\cite{vlog} & BLIP-2 + GRIT & GPT-4 & $\checkmark$ & $\checkmark$ & $1.3$ & $42.3$ & $7.5$ & $2.1$ & $-$ & $-$ \\ 
MM-Vid~\cite{Lin2023mmvid} & GPT-4V & $-$ & $\checkmark$ & $\checkmark$ & $6.1$ & $46.1$ & $9.8$ & $3.8$ & $-$ & $-$ \\ 
MM-Narrator~\cite{Zhang_2024_CVPR} & Azure API + CLIP-L14 & GPT-4V & $\checkmark$ & $\checkmark$ & $9.8$ & $-$ & $12.8$ & $-$ & $7.1$ & $10.9$ \\ 
MM-Narrator~\cite{Zhang_2024_CVPR} & Azure API + CLIP-L14 & GPT-4 & $\checkmark$ & $\checkmark$ & $13.9$ & $49.0$ & $13.4$ & $5.2$ & $6.7$ & $12.8$ \\ 
LLM-AD~\cite{chu2024llmadlargelanguagemodel} & GPT-4V & $-$ & $\checkmark$ & $\checkmark$ & $20.5$ & $-$ & $13.5$ & $-$ & $-$ & $-$ \\ 
AutoAD-Zero~\cite{Xie24b} & GPT-4o & GPT-4o & $\checkmark$ & $\checkmark$ & $25.4$ & $54.3$ & $14.3$ & $8.1$ & $6.7$ & $13.7$ \\ 
\textbf{Ours} & GPT-4o & GPT-4o & $\checkmark$ & $\checkmark$ & $26.9$ & $\mathbf{56.4}$ & $15.0$ & $\mathbf{8.5}$ & $\mathbf{7.4}$ & $15.9$ \\ 
\bottomrule
\end{tabular}
}
\vspace{-0.2cm}
\caption{\textbf{Quantitative comparison on MAD-Eval.} For training-free methods, ``VLM'' and ``LLM'' refer to the models used in separate stages, while for fine-tuned models, they denote the pre-trained components within an end-to-end model.}
\label{tab:quantitative_madeval}
\vspace{-0.2cm}
\end{table*}

\subsection{AD generation -- evaluation of components}
\label{subsec:ad_components}

\vspace{2pt}
\noindent \textbf{Shot-based temporal context.} 
We investigate different setups for leveraging temporal context in AD generation, as shown in~\cref{tab:temporal_context}, leading to the following observations: (i) Expanding the temporal context range noticeably boosts the performance, with gains saturating around ``$2$ shots'' (Exp.\ A, B, and F); (ii) ``Shot number referral'' is the most effective strategy for outlining the current shot. (Exp.\ C, D, and F); (iii) Dynamically sampling the current shots at a higher frame rate boosts AD generation (Exp.\ E and F). The latter two improvements can be attributed to a more efficient focus on the visual content around the targeted AD interval.

\vspace{2pt}
\noindent \textbf{Thread structure injection.}  
After extending the context information with neighbouring shots, we further enhance the VideoLLM's understanding by incorporating thread structures. Note that this guidance is applied only to video sequences exhibiting thread structures. As shown in~\cref{tab:thread_structure_injection}, this additional information improves AD generation performance across both datasets.

\vspace{2pt}
\noindent \textbf{Scale-dependent Stage I factors.} 
\cref{tab:scale_dependent} explores the impact of different Stage I factors on final AD performance. Using shot scales as guidance for Stage I factor formulation (i.e.\ scale-dependent) not only outperforms configurations with single fixed factors but also surpasses the case where all factors are included in Stage I. This could be attributed to that the scale-dependent description contains more relevant and less redundant information, enabling more efficient AD extraction in Stage II.

\subsection{AD generation -- comparison with SotA}
\label{subsec:ad_compare}
\cref{tab:quantitative} provides a comprehensive summary of AD generation performance on CMD-AD and TV-AD, comparing across training-free methods with and without proprietary models, as well as models fine-tuned on human-annotated ADs. Notably, with the same base model setup (Qwen2-VL-7B + LLaMA3-8B), our training-free framework significantly outperforms AutoAD-Zero, primarily due to the usage of temporal context and film grammar information. By incorporating the more powerful GPT-4o models, our performance scales up further, surpassing even existing fine-tuned models.

Regarding the MAD-Eval results, as shown in~\cref{tab:quantitative_madeval}, our framework achieves state-of-the-art performance among all training-free approaches, also demonstrating competitiveness with fine-tuned models.

\begin{figure*}[!htbp]
\centering
\includegraphics[width=0.995\textwidth]{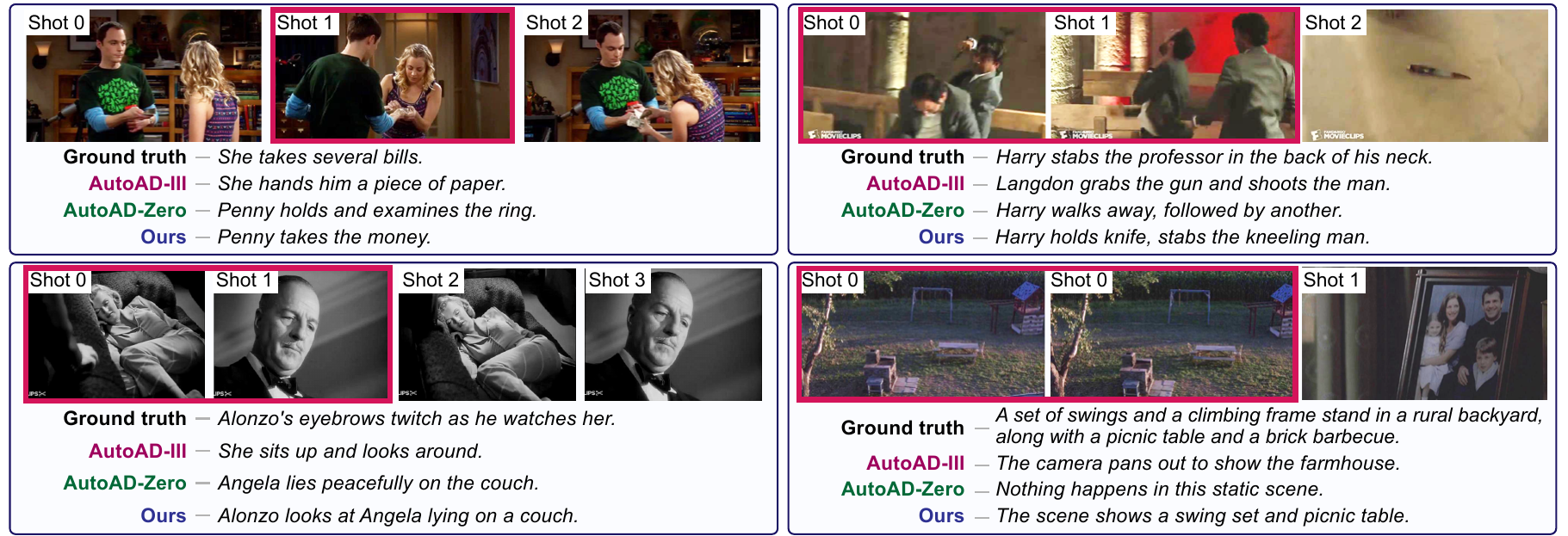}
\vspace{-0.25cm}
    \caption{\small
    \textbf{Qualitative visualisations.} Current shots (corresponding to AD intervals) are outlined by red boxes for illustration purposes only. For simplicity, not all context shots are shown. Training-free methods adopt Qwen2-VL + LLaMA3. Examples are taken from \textit{The Big Bang Theory} (S2E14) (top left), \textit{Inferno} (2016) (top right), \textit{The Asphalt Jungle} (1950) (bottom left), and \textit{Signs} (2002) (bottom right). The top-left example demonstrates the benefits of shot-based temporal context, where the objects (i.e.\ bills) in Penny's hands are not clearly visible within the AD interval (Shot 1), leading to ambiguous or incorrect predictions by AutoAD-Zero. In contrast, our method successfully identifies the objects from the context shot (Shot 2). The top-right example describing the action of \textit{Harry Sims} similarly verifies the effectiveness of incorporating context shots.}
    \vspace{-0.1cm}
    \label{fig:qualitative}
\end{figure*}

\vspace{2pt}
\noindent \textbf{Qualitative visualisations.}
\cref{fig:qualitative} presents several qualitative examples, where the top two cases illustrate how temporal context information aids in identifying key objects.

In the bottom-left example, prior methods fail to associate characters, leading to the omission of the man (\textit{Alonzo}). In contrast, our method recognises the thread structure (i.e.\ \textit{[Shot 0, Shot 2], [Shot 1, Shot 3]}), which guides the correct prediction of the man’s gaze direction towards the lying woman. 

The bottom-right example highlights the effectiveness of scale-dependent Stage I factor formulation. AutoAD-Zero, designed to query characters, actions, interactions, and facial expressions, sometimes overlooks environmental details. Our method, in contrast, correctly identifies the shot as a long shot and instructs the VideoLLM to incorporate environmental context, resulting in more accurate scene descriptions. For more visualisations, please refer to~\cref{supsec:add_vis}.

\begin{table}[t]
\centering
\setlength\tabcolsep{2pt}
\resizebox{0.48\textwidth}{!}{
\begin{tabular}{cccccccc}
\toprule
\multirow{2}[2]{*}{Method} & \multirow{2}[2]{*}{\shortstack{Candidate\\sampling}} & \multicolumn{3}{c}{CMD-AD} & \multicolumn{3}{c}{TV-AD} \\ 
\cmidrule(lr){3-5} \cmidrule(lr){6-8}
 & & CIDEr & CRITIC & Action & CIDEr & CRITIC & Action \\ 
\midrule
\textcolor{gray}{Ours} & \textcolor{gray}{Single AD (Ref.)} & \textcolor{gray}{$26.3$} & \textcolor{gray}{$47.8$} & \textcolor{gray}{$28.4$} & \textcolor{gray}{$31.1$} & \textcolor{gray}{$42.2$} & \textcolor{gray}{$23.9$} \\
\midrule
\multirow{2}{*}{Ours} &  \multirow{2}{*}{\shortstack{Indep. output \\ ($p$\,=\,$0.90; \tau_p$\,=\,$0.6$)}} & \multirow{2}{*}{$33.3$} & \multirow{2}{*}{$49.6$} & \multirow{2}{*}{$32.5$} & \multirow{2}{*}{$41.2$} & \multirow{2}{*}{$44.3$} & \multirow{2}{*}{$28.6$} \\ 
& & & & & & & \\
\multirow{2}{*}{Ours} &  \multirow{2}{*}{\shortstack{Indep. output \\ ($p$\,=\,$0.95; \tau_p$\,=\,$1.5$)}} & \multirow{2}{*}{$37.0$} & \multirow{2}{*}{$\mathbf{50.3}$} & \multirow{2}{*}{$35.1$} & \multirow{2}{*}{$45.5$} & \multirow{2}{*}{$46.4$} & \multirow{2}{*}{$31.5$} \\ 
& & & & & & & \\
\midrule
AutoAD-Zero~\cite{Xie24b} & Joint output & $31.6$ & $46.4$ & $33.8$ & $43.2$ & $46.8$ & $30.1$ \\ 
Ours & Joint output & $\mathbf{38.4}$ & $49.2$ & $\mathbf{35.7}$ & $\mathbf{51.3}$ & $\mathbf{47.3}$ & $\mathbf{31.8}$\\ 
\bottomrule
\end{tabular}
}
\vspace{-0.25cm}
\caption{\small\textbf{Assisted AD generation results.} All methods adopt Qwen2-VL + LLaMA3-8B as base models. The first row provides single AD generation results as references (labelled in gray), the rest rows report the performance of one selected AD out of five candidates. ``Indep. output'' denotes five random independent Stage II runs, with $p$ as the hyperparameter for top-p (nucleus) sampling and $\tau_p$ as the sampling temperature. ``Joint output'' generates five ADs simultaneously in a single run. 
}
\label{tab:assistant}
\vspace{-0.3cm}
\end{table}
\subsection{Assisted AD generation}
\label{subsec:ad_assist}
The subjective nature of AD sets a practical limit on metric scores, lower than the theoretical maximum, because human annotators often provide different but valid descriptions.
Therefore, beyond enforcing generating a \emph{single} AD sentence, we also consider employing our framework as an \textit{assistant} to produce \textit{multiple} candidate AD sentences.

To standardise such a protocol, we consider five candidate ADs generated by an \textit{assistant} and employ an \textit{expert} to select the best one. To effectively benchmark performance against existing GT ADs, we define the \textit{``expert''} as an automatic selection mechanism that chooses the candidate with the highest average CIDEr and action score.

To develop an AD generation assistant, we fix the Stage I dense descriptions and explore generating multiple candidates in Stage II. This can be achieved by either running Stage II independently five times (termed the ``independent output'' setup) or generating five AD outputs simultaneously within a single run (termed the ``joint output'' setup).
As observed in~\cref{tab:assistant}, the assistant-based setup significantly improves upon the single AD performance, highlighting the potential of training-free methods in effectively capturing the desired content for AD generation. Within the ``independent output'' setup, increasing the randomness of sampling (i.e.\ higher $p$ and $\tau_p$) enhances the quality of the selected AD, owing to greater candidate diversity. Meanwhile, the ``joint output'' setup achieves superior performance, which could be attributed to reduced information redundancy across the simultaneously generated ADs. For additional visualisations, discussions, and detailed text prompts, please refer to~\cref{supsec:add_vis}.

\vspace{0.25cm}
\section{Discussion -- summary and limitations}
\label{sec:conclusion}
\vspace{0.1cm}
We have demonstrated the benefit of shot-based context and film grammar awareness in AD generation --  our training-free two-stage framework achieves state-of-the-art performance among all training-free counterparts, even surpassing fine-tuned models on multiple benchmarks. 

The current framework has two main limitations: (i) the performance depends on the base VideoLLM, which may occasionally hallucinate details inconsistent with the visual content; and
(ii) story-level context is not incorporated into the AD generation process.  
These limitations could potentially be addressed in future work
by improving visual grounding and extending the visual and textual context to include the plot.

\clearpage
\clearpage
\paragraph{Acknowledgements.}
This research is supported by the UK EPSRC Programme Grant Visual AI (EP/T028572/1), a Royal Society Research Professorship RSRP$\backslash$R$\backslash$241003, a Clarendon Scholarship, and ANR-21-CE23-0003-01 CorVis.

{
    \small
    \bibliographystyle{ieeenat_fullname}
    \bibliography{main,vgg_local}
}

\clearpage

\renewcommand{\thefigure}{A\arabic{figure}} 
\setcounter{figure}{0} 
\renewcommand{\thetable}{A\arabic{table}}
\setcounter{table}{0} 

\maketitlesupplementary
\appendix

\noindent This appendix is organised as follows:  
\vspace{0.1cm}
\begin{itemize}  
    \item \textbf{Implementation details:} In~\cref{supsec:imple}, we provide additional details on film grammar prediction, action score evaluation settings, and the exact text instructions.  
    \item \textbf{Evaluation metrics for AD generation:} In~\cref{supsec:metric}, we elaborate on the key metrics used to measure AD quality.  
    \item \textbf{Additional experimental results:} In~\cref{supsec:add_abla}, we provide additional results for AD generation.
    \item \textbf{Human alignment with action scores:} In~\cref{supsec:action_score}, we provide a thorough description of the human agreement study for action scores, including curated test sets, a correlation analysis, and an inter-rater agreement study.  
    \item \textbf{Qualitative visualisations:} \cref{supsec:add_vis} includes more detailed visualisations, along with an in-depth analysis of failure cases and assisted AD generation.  
\end{itemize}

\section{Implementation details}
\label{supsec:imple}

\vspace{3pt} \noindent \textbf{Thread structure prediction setting.}  
To predict thread structure in a zero-shot manner, we resize video frames to $224$p and employ DINOv2~\cite{oquab2024dinov} ViT-g/14 w.\ reg to extract spatial features. During matching score prediction, we consider a $5 \times 5$ mask neighbourhood for cost volume computation and set the softmax temperature to \( \tau = 0.1 \). To determine the relationship between each pair of shots, we apply a threshold of \( \epsilon = 0.3 \) to the matching score \( s^{i,j} \).

\vspace{3pt} \noindent \textbf{Shot scale classification setting.}  
For shot scale classification, we fine-tune the last $6$ layers of the DINOv2~\cite{oquab2024dinov} ViT-B/14 on the MovieShots training set. During evaluation, we use the middle frame of each shot as input.

The averaged shot scale of the current shots (i.e.\ the effective shot scale $S_{\text{eff}}$) is used to guide the incorporation of additional factors in Stage I prompts. Specifically, 
\begin{align}
    \text{Stage I factor +=} \begin{cases}
    \text{Facial expression}, & \text{if $S_{\text{eff}} \leq 1.5$}\\ \nonumber
    \text{Key object}, & \text{if $2 \leq S_{\text{eff}} \leq 3$}\\ \nonumber
    \text{Environment}, & \text{if $S_{\text{eff}} \geq 3.5$}\\ \nonumber
    \text{None} , & \text{otherwise}
  \end{cases}
\end{align}

\vspace{3pt} \noindent \textbf{Action score evaluation setting.}  
To obtain character-free GT action sentences, we employ LLaMA3.1-70B~\cite{grattafiori2024llama3herdmodels} for pre-processing in two steps:
(i) \textit{Character information removal:} Character names are replaced with appropriate pronouns using the LLM with the prompt provided in~\cref{supalg:char_info_removal};
(ii) \textit{Action sentence extraction:} Each AD sentence is split into subsentences, each containing a single action. To achieve this, the LLM is prompted with instructions in~\cref{supalg:action}.

During hierarchical prediction parsing, we use spaCy\footnote{\url{https://spacy.io/models/en}} to extract action phrases and corresponding verb lemmas from predicted sentences. 

During similarity score computation, to extract sentence embeddings, we apply {gte-Qwen2-7B-instruct}~\cite{li2023towards}, which supports optional text prompt input as guidance. We set the prompt to: \textit{``Retrieve relevant passages that involve similar actions, with particular focus on the verbs.''}, further emphasising actions and verbs during similarity matching.

Additionally, to establish the LLM-based baseline, we define evaluation criteria as outlined in~\cref{supalg:llm_based} and use them to prompt LLaMA-3.1-70B and GPT-4o.

\vspace{3pt}
\noindent \textbf{GPT-4o setup for AD generation.} 
For both stages, we use \texttt{gpt-4o-2024-08-06}~\cite{openai2024gpt4technicalreport} as the base model. For visual token extraction, the ``detail'' parameter is set to ``low''.

\vspace{3pt}  
\noindent \textbf{Text instructions for AD generation.}  
The prompts for AD generation are provided in~\cref{supalg:stage1,supalg:stage2} for Stage I and Stage II, respectively. The Stage I prompt is designed for both Qwen2-VL and GPT-4o, while the Stage II prompt is tailored for LLaMA3 and GPT-4o.

\begin{table*}[t]
\centering
\setlength\tabcolsep{8pt}
\resizebox{0.975\textwidth}{!}{
\begin{tabular}{cccccccccc}
\toprule
\multirow{3}[2]{*}{\shortstack{Stage I\\VideoLLM}} & \multirow{3}[2]{*}{\shortstack{Stage II\\LLM}} & \multicolumn{4}{c}{CMD-AD} & \multicolumn{4}{c}{TV-AD} \\ 
\cmidrule(lr){3-6} \cmidrule(lr){7-10}
&  & \multirow{2}{*}{CIDEr} & \multirow{2}{*}{CRITIC} & \multirow{2}{*}{\shortstack{Action\\(Stage I)}} & \multirow{2}{*}{\shortstack{Action\\(Stage II)}}  & \multirow{2}{*}{CIDEr} & \multirow{2}{*}{CRITIC} & \multirow{2}{*}{\shortstack{Action\\(Stage I)}} & \multirow{2}{*}{\shortstack{Action\\(Stage II)}}  \\ 
& & & & & & & & \\
\midrule
Qwen2.5-VL-7B~\cite{Qwen2.5-VL} & LLaMA3-8B~\cite{grattafiori2024llama3herdmodels}  & $24.1$  & $\mathbf{49.7}$  & $35.5$ & $27.2$  & $26.5$  & $\mathbf{43.6}$ & $36.4$ & $23.8$\\ 
VideoLLaMA3-7B~\cite{damonlpsg2025videollama3} & LLaMA3-8B~\cite{grattafiori2024llama3herdmodels}  & $22.1$  & $45.1$  & $34.2$ & $24.8$  & $26.8$  & $41.3$ & $\mathbf{39.9}$ & $23.8$\\ 
InternVL2.5-8B~\cite{chen2024expanding} & LLaMA3-8B~\cite{grattafiori2024llama3herdmodels}  & $24.0$  & $46.0$  & $35.0$ & $28.1$  & $28.3$  & $41.2$ & $36.0$ & $\mathbf{24.2}$\\ 
Qwen2-VL-7B~\cite{Qwen2VL} & LLaMA3-8B~\cite{grattafiori2024llama3herdmodels}  & $\mathbf{26.3}$  & $47.8$  & $\mathbf{38.2}$ & $\mathbf{28.4}$  & $\mathbf{31.1}$  & $42.2$ & $38.2$ & $23.9$\\ 
\midrule
Qwen2-VL-7B~\cite{Qwen2VL} & Gemma3-12B~\cite{gemmateam2025gemma3}  & $\mathbf{26.4}$  & $45.9$  & $-$ & $\mathbf{30.6}$  & $30.0$  & $43.2$ & $-$ & $\mathbf{24.7}$\\ 
Qwen2-VL-7B~\cite{Qwen2VL} & Qwen3-8B~\cite{qwen3}  & $26.3$  & $45.2$  & $-$ & $28.4$  & $30.7$  & $\mathbf{43.4}$ & $-$ & $\mathbf{24.7}$\\ 
Qwen2-VL-7B~\cite{Qwen2VL} & LLaMA3-8B~\cite{grattafiori2024llama3herdmodels}  & $26.3$  & $\mathbf{47.8}$  & $-$ & $28.4$  & $\mathbf{31.1}$  & $42.2$ & $-$ & $23.9$\\ 
\addlinespace[0.1em]
\arrayrulecolor{gray}\hdashline
\addlinespace[0.3em]
\textcolor{gray}{GPT-4o}~\cite{openai2024gpt4technicalreport} & \textcolor{gray}{GPT-4o}~\cite{openai2024gpt4technicalreport}  & \textcolor{gray}{$26.1$}  & \textcolor{gray}{$49.1$}  & \textcolor{gray}{$40.2$} & \textcolor{gray}{$32.5$}  & \textcolor{gray}{$34.2$}  & \textcolor{gray}{$46.5$}  & \textcolor{gray}{$41.0$} & \textcolor{gray}{$27.4$}\\ 
\bottomrule
\end{tabular}
}
\vspace{-0.2cm}
\caption{\textbf{Different open-source VideoLLMs (for Stage I) and LLMs (for Stage II).} As a reference, the last row reports results using the proprietary GPT-4o model. Note, we additionally report action scores for predicted Stage I description, whereas other metrics, including CIDEr, CRITIC, and Action (Stage II), measure the Stage II AD quality.}
\label{suptab:videollms}
\vspace{-0.2cm}
\end{table*}

\section{Evaluation metrics for AD generation}
\label{supsec:metric}

\noindent \textbf{CIDEr}~\cite{Vedantam_2015_CVPR} 
measures text similarity by computing a weighted word-matching score, emphasising n-gram overlap while accounting for term frequency and importance through TF-IDF~\cite{Robertson2004} weighting.

\vspace{2pt}  
\noindent \textbf{Recall@k/N}~\cite{Han23} is a retrieval-based metric that evaluates whether predicted texts can be distinguished from their temporal neighbours. Specifically, for each predicted AD, it checks whether the AD can be retrieved at a top-k position within a neighbourhood of N ADs. Following prior work~\cite{Han24, distinctad}, we report Recall@1/5 on CMD-AD and TV-AD, and Recall@5/16 on MAD-Eval. 

\vspace{2pt}  
\noindent \textbf{LLM-AD-Eval}~\cite{Han24} employs LLM agents (LLaMA2-7B~\cite{touvron2023llama2openfoundation} $|$ LLaMA3-8B~\cite{grattafiori2024llama3herdmodels}) as evaluators to compare ground truth ADs with predictions, generating a matching score ranging from 1 (lowest) to 5 (highest).

\vspace{2pt}  
\noindent \textbf{CRITIC}~\cite{Han24} measures the accuracy of character names in predicted ADs. It first resolves character ambiguity in GT ADs by applying a coreference model to replace pronouns with corresponding character names. During evaluation, the intersection-over-union (IoU) of predicted and ground truth character names is computed.

\vspace{2pt}  
\noindent \textbf{Action Score,} as detailed in~\cref{sec:action_score}, evaluates the quality of predicted actions (i.e. verbs, object nouns, etc.) while minimising the influence of character name variations. Throughout this paper, unless otherwise specified, we use the action score to assess Stage II AD outputs, i.e. ``Action'' refers to ``Action (Stage II)''.

\vspace{3pt}  
\noindent For MAD-Eval, we additionally report the performance on conventional metrics including ROUGE-L~\cite{lin-2004-rouge}, SPICE~\cite{spice2016}, METEOR~\cite{banerjee-lavie-2005-meteor}, and BLEU-1~\cite{bleu_score}.

\begin{table}[t]
\centering
\setlength\tabcolsep{8pt}
\resizebox{0.475\textwidth}{!}{
\begin{tabular}{ccccccc}
\toprule
\multirow{2}[2]{*}{\shortstack{Thread\\structure}} & \multirow{2}[2]{*}{\shortstack{Stage I \\VideoLLM}} & \multicolumn{3}{c}{TV-AD subset} \\ 
\cmidrule(lr){3-5} 
& & CIDEr & CRITIC & Action  \\ 
\midrule
\xmark & Qwen2.5-VL-7B~\cite{Qwen2.5-VL} & $26.7$ & $42.8$ & $23.7$  \\ 
$\checkmark$ & Qwen2.5-VL-7B~\cite{Qwen2.5-VL} & $27.9$\greenarrow{1.2} & $43.4$\greenarrow{0.6} & $23.7$\graytext{0.0} \\ 
\addlinespace[0.1em]
\arrayrulecolor{gray}\hdashline
\addlinespace[0.2em]
\xmark & VideoLLaMA3-7B~\cite{damonlpsg2025videollama3} & $24.9$ & $42.1$ & $21.9$ \\ 
$\checkmark$ & VideoLLaMA3-7B~\cite{damonlpsg2025videollama3} & $25.5$\greenarrow{0.6} & $41.1$\redarrow{1.0} & $22.7$\greenarrow{0.8} \\ 
\addlinespace[0.1em]
\arrayrulecolor{gray}\hdashline
\addlinespace[0.2em]
\xmark & InternVL2.5-8B~\cite{chen2024expanding} & $25.9$ & $40.3$ & $21.8$  \\ 
$\checkmark$ & InternVL2.5-8B~\cite{chen2024expanding} & $27.1$\greenarrow{1.2} & $41.9$\greenarrow{1.6} & $23.1$\greenarrow{1.3} \\
\addlinespace[0.1em]
\arrayrulecolor{gray}\hdashline
\addlinespace[0.2em]
\xmark & Qwen2-VL-7B~\cite{Qwen2VL} & $28.8$ & $42.0$ & $22.6$  \\ 
$\checkmark$ & Qwen2-VL-7B~\cite{Qwen2VL}  & $30.7$\greenarrow{1.9} & $42.7$\greenarrow{0.7} & $22.9$\greenarrow{0.3}  \\
\bottomrule
\end{tabular}
}
\vspace{-0.2cm}
\caption{\textbf{Thread structure injection for different open-source VideoLLMs.} The base model for Stage II is LLaMA3-8B. Thread structure information is injected only into subsets predicted to exhibit thread structures ($\sim\!60\%$ in TV-AD).}
\vspace{-0.2cm}
\label{suptab:thread_videollm}
\end{table}

\section{Additional experimental results}
\label{supsec:add_abla}

\vspace{3pt}
\noindent \textbf{Different VideoLLMs for Stage I.} 
\cref{suptab:videollms} compares our AD generation performance using different open-source VideoLLMs in Stage I, validating our choice of Qwen2-VL-7B as the default model.  

Beyond the Stage II action scores presented in the main text, we also report Stage I action scores as a direct indicator of dense description performance. 
In general, Stage I action scores are noticeably higher than their Stage II counterparts, suggesting that some ground truth actions are captured in dense descriptions but are not selected for the final AD outputs. This further supports the validity of our assisted AD generation protocol, where multiple candidate ADs with different actions are extracted from dense descriptions and await further selection.

\vspace{3pt}
\noindent \textbf{Different LLMs for Stage II.} 
\cref{suptab:videollms} also compares different options for the Stage II LLM. Overall, the choice of LLM has a relatively minor impact compared to the Stage I VideoLLM. The default LLaMA3-8B achieves overall strong performance.

\vspace{3pt}
\noindent \textbf{Thread structure injection for different VideoLLMs.}  
We investigate how injecting thread structure into different open-source VideoLLMs affects AD generation. Specifically, we evaluate performance on TV-AD, which contains a large proportion of thread-structured video clips. As shown in~\cref{suptab:thread_videollm}, incorporating thread information leads to general performance boosts across various VideoLLMs.

\vspace{3pt}
\noindent \textbf{Repeated AD generation.} \cref{tab:repeated_experiments} reports results from five independent runs of our two-stage AD generation pipeline, each adopting a different random seed.
The results are largely consistent across runs, indicating the stability of the AD generation process. In particular, the CRITIC results exhibit the highest variance, followed by CIDEr, while the action score remains relatively stable across repeated experiments. 

\begin{table}[t]
\centering
\setlength\tabcolsep{7pt}
\resizebox{0.48\textwidth}{!}{
\begin{tabular}{ccccccc}
\toprule
\multirow{2}[2]{*}{Exp.} & \multicolumn{3}{c}{CMD-AD} & \multicolumn{3}{c}{TV-AD} \\ 
\cmidrule(lr){2-4} \cmidrule(lr){5-7}
 & CIDEr & CRITIC & Action & CIDEr & CRITIC & Action \\ 
\midrule
1 & $26.3$ & $47.8$ & $28.4$ & $31.1$ & $42.2$ & $23.9$ \\ 
2 & $26.9$ & $47.7$ & $28.4$ & $30.7$ & $41.9$ & $23.8$ \\ 
3 & $26.4$ & $48.3$ & $28.3$ & $30.5$ & $43.0$ & $23.7$ \\ 
4 & $26.3$ & $47.3$ & $28.5$ & $30.8$ & $42.0$ & $23.5$ \\ 
5 & $26.5$ & $48.3$ & $28.4$ & $31.5$ & $41.9$ & $23.7$ \\ 
\midrule
Mean & $26.5$ & $47.9$ & $28.4$ & $30.9$ & $42.2$ & $23.7$ \\ 
STD & $0.2$ & $0.4$ & $0.1$ & $0.4$ & $0.5$ & $0.1$\\ 
\bottomrule
\end{tabular}
}
\vspace{-0.2cm}
\caption{\textbf{Repeated (multi-run) experiments.} Results shown are from five independent runs (Stage I + Stage II) using different random seeds.}
\label{tab:repeated_experiments}
\end{table}

\begin{table}[t]
\centering
\setlength\tabcolsep{5pt}
\resizebox{0.47\textwidth}{!}{
\begin{tabular}{cccccc}
\toprule
 \multirow{2}{*}{Method} & \multirow{2}{*}{\shortstack{Shot \\ Partition}} &  \multirow{2}{*}{\shortstack{Film \\ Grammar}} & \multirow{2}{*}{\shortstack{Stage I \\ VideoLLM}}  & \multirow{2}{*}{\shortstack{\;Stage II\; \\ LLM}} & \multirow{2}{*}{\;Total\;} \\ 
 & & & & & \\ 
\midrule
AutoAD-Zero & $-$ & $-$ & $2.18$s & $0.64$s & $2.82$s \\  
Ours & $0.09$s &  $0.12$s & $2.82$s & $0.72$s & $3.75$s \\ 
\bottomrule
\end{tabular}
}
\vspace{-0.15cm}
\caption{\textbf{Inference time analysis.}}
\label{suptab:inference_time}
\vspace{-0.2cm}
\end{table}

\vspace{3pt}  
\noindent \textbf{Inference time analysis.}
We report the inference time \textit{per AD} in~\cref{suptab:inference_time}. Our added components (e.g.\ shot partitioning, film grammar prediction) incur minimal overhead. The main cost increase arises from sampling more contextual frames for Stage I inputs. Overall, our method maintains reasonable efficiency—given averaged input clip duration (including contextual shots) is $19.82$s—while delivering clear performance gains.

\begin{figure}[t]
\centering
\includegraphics[width=0.485\textwidth]{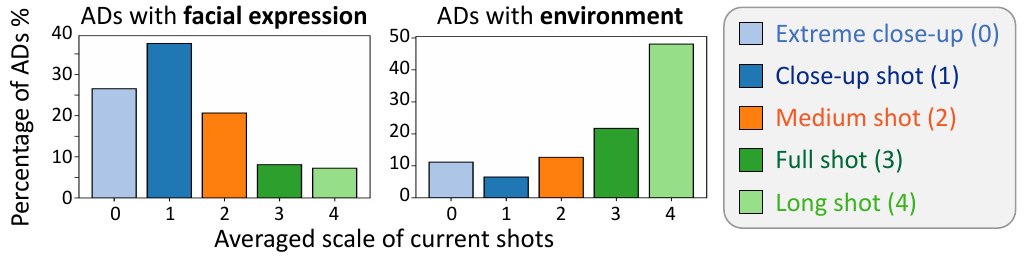}
\vspace{-0.65cm}
\caption{\small
\textbf{Correlations between descriptive factors in ADs and averaged shot scales.}
\label{supfig:factor_vs_scale}
}
\vspace{-0.4cm}
\end{figure}

\vspace{3pt}  
\noindent \textbf{Investigation on factor--scale correlations.} To verify our assumptions about the strong correlations between descriptive factors in ADs and shot scales (as mentioned in~\cref{subsec:shot}), we used GPT-4o to extract key elements (e.g.\ environments) mentioned in GT ADs from $100$ movies, and analysed their correlation with shot scales. As shown in~\cref{supfig:factor_vs_scale}, facial expressions predominantly occur in (extreme) close-up shots, while environmental cues align with full/long shots. These observations align with the assumption and generalise across real-world movies.

\begin{table*}[t!]
\centering
\small
\setlength\tabcolsep{6pt}
\resizebox{0.985\textwidth}{!}{
\begin{tabular}{>{\centering\arraybackslash}p{0.04\textwidth} p{0.2\textwidth} p{0.2\textwidth} p{0.56\textwidth} }
\toprule
Score  & Note &  GT action & Prediction  \\ 
\midrule
0  &  \textbf{Unrelated}: \newline  Completely irrelevant in actions, objects, environments, etc. & \textit{He indicates that he was just joking.}  & \textit{The video begins with a man sitting under an umbrella in the rain, wearing a dark jacket with red accents and a blue shirt underneath. The scene transitions to show him lying on the ground, wet from the rain, with his arms outstretched and his eyes closed. Another person, dressed in a white shirt and black vest, reaches towards him from above.}  \\
\cmidrule(lr){3-4}
&  & \textit{His new companion cuddles back up to him.} & \textit{The flames start to die down.} \\
\midrule
1  & \textbf{Loosely related}: \newline  Semantically or causally \newline relevant actions; \newline 
 Similar objects, \newline environments, etc.  & \textit{He points the remote at the word's main menu.}  & \textit{1. Main characters: Morty and Michael Newman. 2. Actions: Both characters are standing in a room with futuristic digital screens and symbols around them. They appear to be engaged in an activity involving the screens, possibly \textbf{interacting with them or using devices}. 3. Character-character interactions: The two characters seem to be focused on the screens, possibly discussing or collaborating on something related to the content displayed.}  \\
\cmidrule(lr){3-4}
&  & \textit{They step away from the window.}  & \textit{The four of them \textbf{leave the cabin} and go to their horses tethered to a rail at the side.}  \\
\midrule
2  & \textbf{Similar meaning}: \newline  Similar verb, objects, \newline  environments, etc.  & \textit{He dives into an Olympic-sized swimming pool.} & \textit{The movie clip shows a young boy running out of a house wearing only a diaper. He then jumps over a fence and runs across a lawn. The boy continues running and \textbf{jumps into a swimming pool}. Throughout the video, the boy's movements are energetic and playful, and he appears to be having fun.} \\
\cmidrule(lr){3-4}
&  & \textit{She carries a laundry basket outside the house.} & \textit{Caroline \textbf{brings some washing} into the garden.}  \\
\midrule
3  & \textbf{Exactly same match:}  \newline  Exactly the same verb;  \newline  Same/similar objects, \newline  environments, etc. & \textit{He lights a cigar.} & \textit{In the movie clip, a man is seen sitting at a table, while another man enters the room and approaches him. The second man takes out a \textbf{cigarette} and \textbf{lights it}, while the first man watches him. The scene is set in a dimly lit room with a table and chairs in the foreground, and a window in the background.}\\
\cmidrule(lr){3-4}
&  & \textit{She stares glumly at the night sky.}  & \textit{At the palace, Jasmine wanders out into her balcony and \textbf{stares up at the stars}.} \\
\bottomrule
\end{tabular}
}
\vspace{-0.3cm}
\caption{\textbf{Example of human-annotated scores} assessing whether the ground truth (GT) action is accurately captured in the predictions. For each score, examples of a paragraph prediction and a sentence prediction are provided.}
\label{suptab:score_example}
\end{table*}

\begin{figure*}[htbp]
\centering
\includegraphics[width=0.99\textwidth]{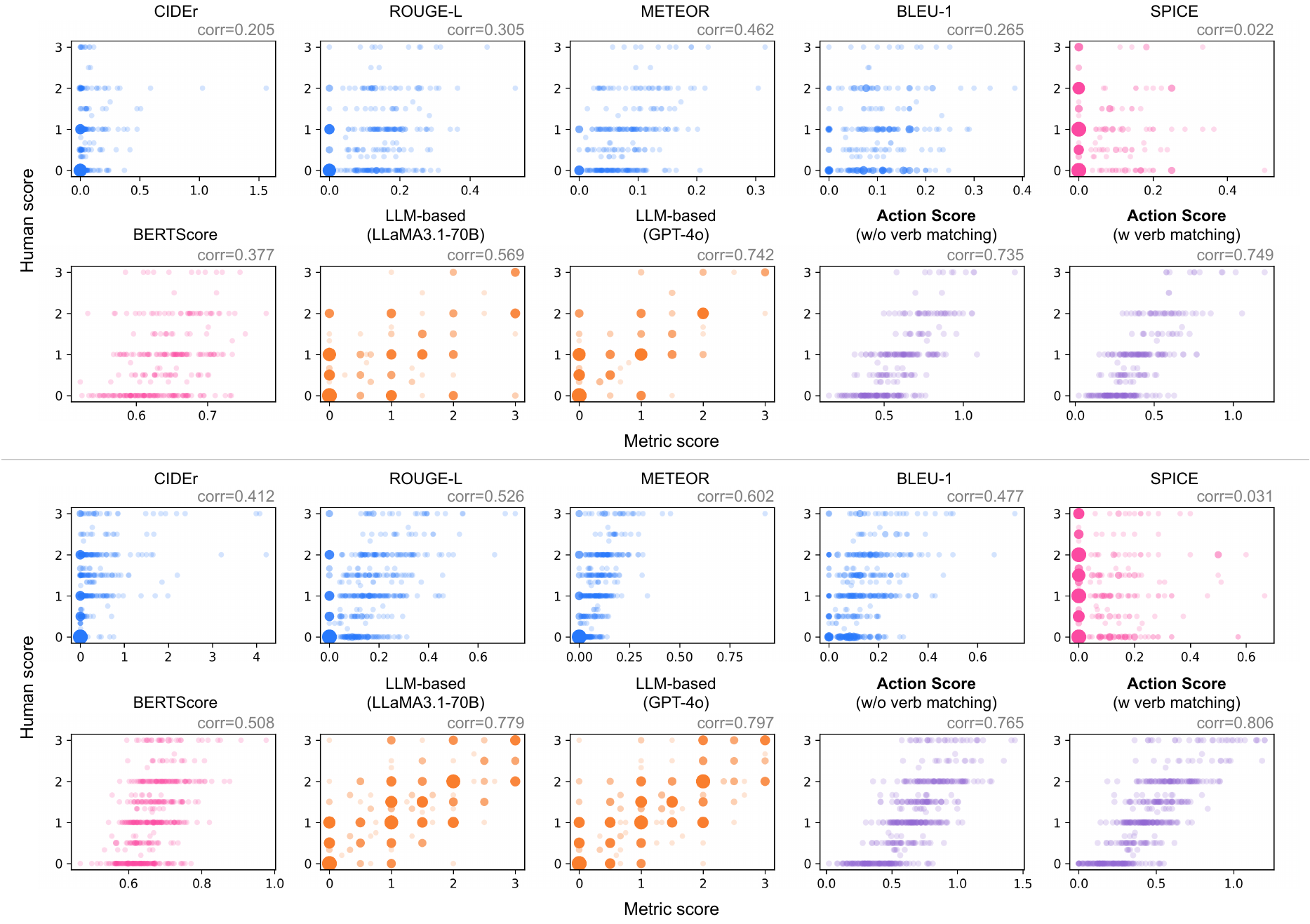}
\vspace{-0.3cm}
    \caption{
    \textbf{Human agreement of different metrics for action evaluation,} where human-annotated scores are compared with metric scores. These scores measure the quality of the Stage I description paragraph (top) or the Stage II output AD (bottom). We consider various metrics, including word-matching-based metrics (blue), semantic-based metrics (pink), LLM-based metrics (orange), and our proposed action scores (purple). The Pearson correlation between human and metric scoring is reported. Within the scatter plots, we use colour depth and marker size to indicate density -- larger and darker markers represent more data points at a single position. Zoom in for a clearer view.
    }
    \label{supfig:scatter}
\end{figure*}

\section{Human alignment with action scores}
\label{supsec:action_score}
To assess how the proposed action score aligns with human knowledge, we curate a test set that measures the correlation between action scores and human scoring.

\vspace{3pt}
\noindent \textbf{Scoring criteria.} The human-annotated scores measure whether the ground truth (GT) action is described in the descriptions, ranging from $0$ to $3$ based on the following scoring criteria:
\begin{itemize}
    \item Score 0 - \textit{GT action is unrelated to any action in PD}
    \item Score 1 - \textit{GT action is loosely related to an action in PD}
    \item Score 2 - \textit{GT action is similar in meaning to an action in PD}
    \item Score 3 - \textit{GT action exactly matches with an action in PD, using the same verb}
\end{itemize}
where PD stands for Predicted Description. Additional guidelines and scoring examples are provided in~\cref{suptab:score_example}.

\vspace{3pt}
\noindent \textbf{Test Set Formulation.}  
We construct two test sets, namely the ``paragraph set'' and the ``sentence set,'' corresponding to the scoring of (Stage I) dense descriptions and (Stage II) AD sentences, respectively.  

The paragraph set consists of $300$ ground truth (GT) ADs, each paired with a predicted paragraph. In total, around $460$ character-free GT actions are extracted, with each action-paragraph pair manually annotated by five workers using the $0$–$3$ scoring scale, as described in the previous section. For ADs containing multiple GT actions, the final human score is obtained by averaging the manually assigned scores across different actions, meaning the resultant score may not always be an integer.  

The sentence set contains $500$ GT ADs with approximately $890$ actions. Given the video clip described by the GT AD, instead of generating AD predictions from a VLM, we use a human-narrated AD from alternative sources for the same clip as the prediction. Similarly, the final human score is computed by averaging the scores across different actions within each GT AD.

\begin{table}[t]
\centering
\setlength\tabcolsep{4pt}
\resizebox{0.48\textwidth}{!}{
\begin{tabular}{ccc|cccc}
\toprule
tIoU & \#movies & \#AD pairs & CIDEr & R@1/5 & Action & LLM-AD-Eval \\ 
\midrule
$0.8$ & $315$ & $4447$ & $61.5$ & $71.2$ & $45.6$ & $3.04 \;|\; 3.24$  \\ 
$0.9$ & $267$ & $999$ & $69.8$ & $80.4$ & $47.6$ & $3.53 \;|\; 3.34$  \\ 
$0.95$ & $148$ & $229$ & $73.9$ & $-$ & $47.8$  & $3.57 \;|\; 3.45$  \\ 
\bottomrule
\end{tabular}
}
\vspace{-0.2cm}
\caption{\textbf{Inter-rater analysis on CMD-AD,} where two versions of human-annotated ADs for the same movie clip are compared under different temporal intersection-over-union (tIoU) thresholds.}
\vspace{-0.3cm}
\label{tab:inter-rater}
\end{table}

\vspace{3pt}
\noindent \textbf{Correlation between human scoring and metrics.} 
We plot the human-annotated scores against the scores reported by each metric, as shown in~\cref{supfig:scatter}. Most conventional metrics (blue and pink) fail to align with human evaluations of action predictions. In contrast, both LLM-based metrics and our action scores effectively assess the quality of action predictions in AD sentences (\cref{supfig:scatter}, bottom).  

However, when evaluating longer paragraphs against the GT action, LLM-based metrics struggle, whereas action scores maintain a high correlation with human judgments (\cref{supfig:scatter}, top). Quantitative results on human-metric correlations are provided in~\cref{tab:action_scores}.

\vspace{3pt} \noindent \textbf{Inter-rater analysis for action score.}
To obtain an upper bound on action scores for predicted AD sentences, we measure the agreement between two versions of human-annotated ADs for the same movie clip at different temporal IoUs (i.e.\ inter-rater agreement~\cite{Han24}). As shown in~\cref{tab:inter-rater}, the action score increases monotonically as the temporal IoU increases.

\section{Qualitative visualisations}
\label{supsec:add_vis}

\begin{figure*}[htbp]
\centering
\includegraphics[width=0.99\textwidth]{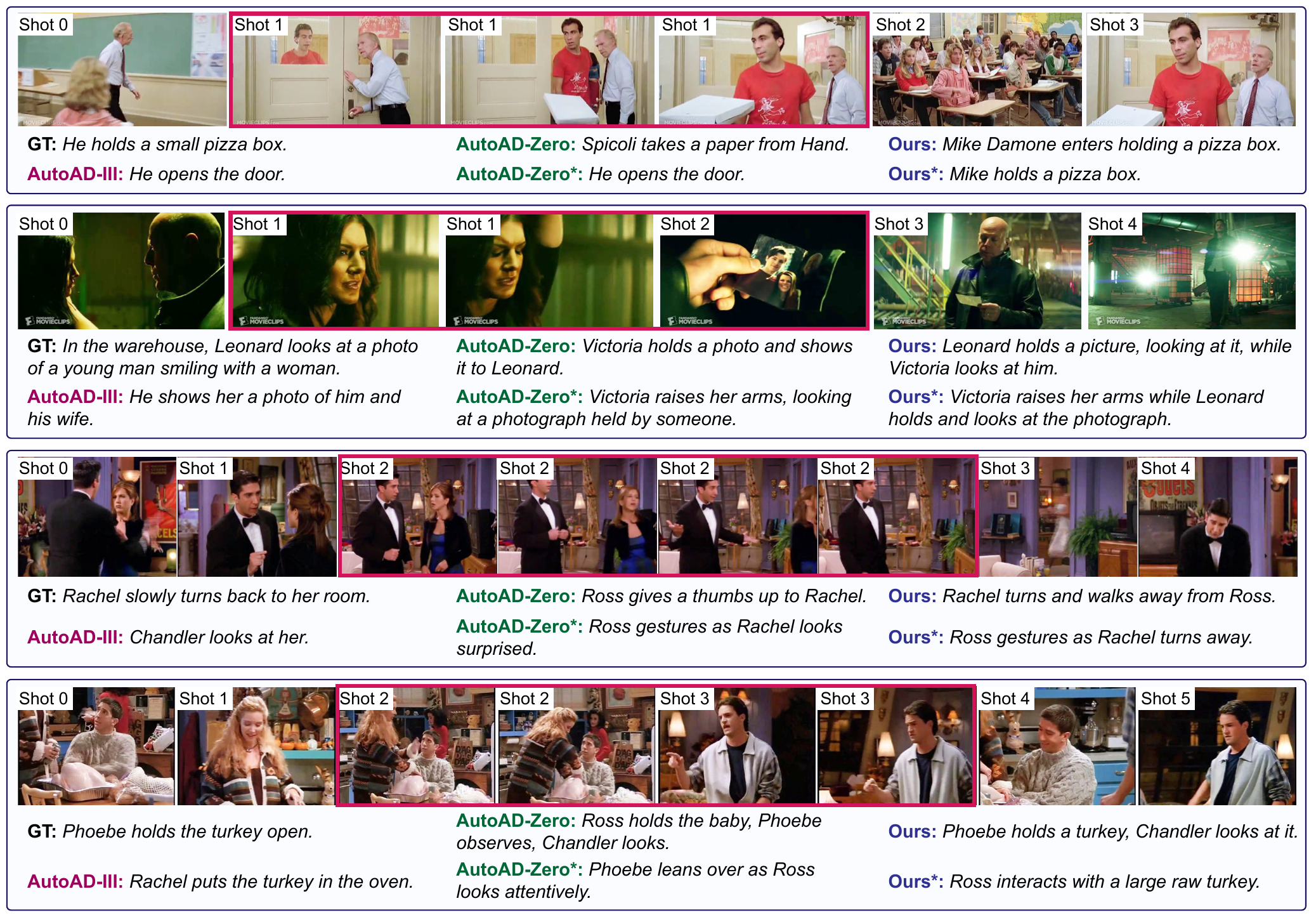}
\vspace{-0.2cm}
    \caption{
    \textbf{More qualitative visualisations.} Current shots (corresponding to AD intervals) are outlined by red boxes for illustration purposes only. Training-free methods with ``*'' adopt GPT-4o (otherwise Qwen2-VL + LLaMA3). Examples from top to bottom are taken from \textit{Fast Times at Ridgemont High} (1982), \textit{Extraction} (2015), \textit{Friends} (S3E2), and \textit{Friends} (S1E9), respectively.}
    \label{supfig:sup_vis}
\end{figure*}

\cref{supfig:sup_vis} presents additional visualisations for CMD-AD and TV-AD, comparing our method against other approaches.

\begin{figure*}[htbp]
\centering
\includegraphics[width=0.99\textwidth]{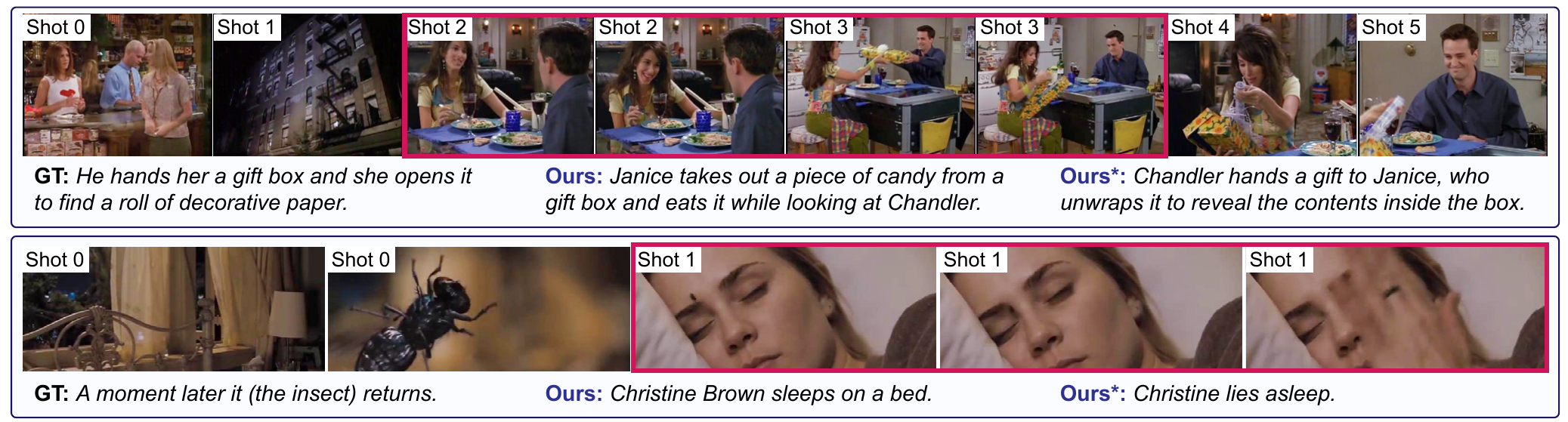}
\vspace{-0.2cm}
    \caption{
    \textbf{Failure case visualisations.} Current shots (corresponding to AD intervals) are outlined by red boxes for illustration purposes only. Training-free methods with ``*'' adopt GPT-4o (otherwise Qwen2-VL + LLaMA3). Examples from top to bottom are taken from \textit{Friends} (S3E4), and \textit{Drag Me to Hell} (2009), respectively.}
    \label{supfig:sup_vis_failure}
\end{figure*}

\vspace{3pt}
\noindent \textbf{Example failure cases} are illustrated in~\cref{supfig:sup_vis_failure}. The top example highlights a hallucination issue in the prediction by Qwen2-VL + LLaMA3 (Ours), where the VideoLLM model misidentifies a ``gift'' as ``candy'' and infers an incorrect action of ``eating,'' which does not occur. When a stronger base model is used (Ours* with GPT-4o), this issue is mitigated.

Additionally, the current method struggles to incorporate broader (story-level) contextual understanding, as demonstrated in the bottom example of~\cref{supfig:sup_vis_failure}. Specifically, the model fails to describe the return of an insect and instead focuses on the sleeping woman. This limitation could potentially be addressed by incorporating more abstract information, which could be extracted from existing movie plots or summarised from a longer temporal context.

\begin{figure*}[htbp]
\centering
\includegraphics[width=0.99\textwidth]{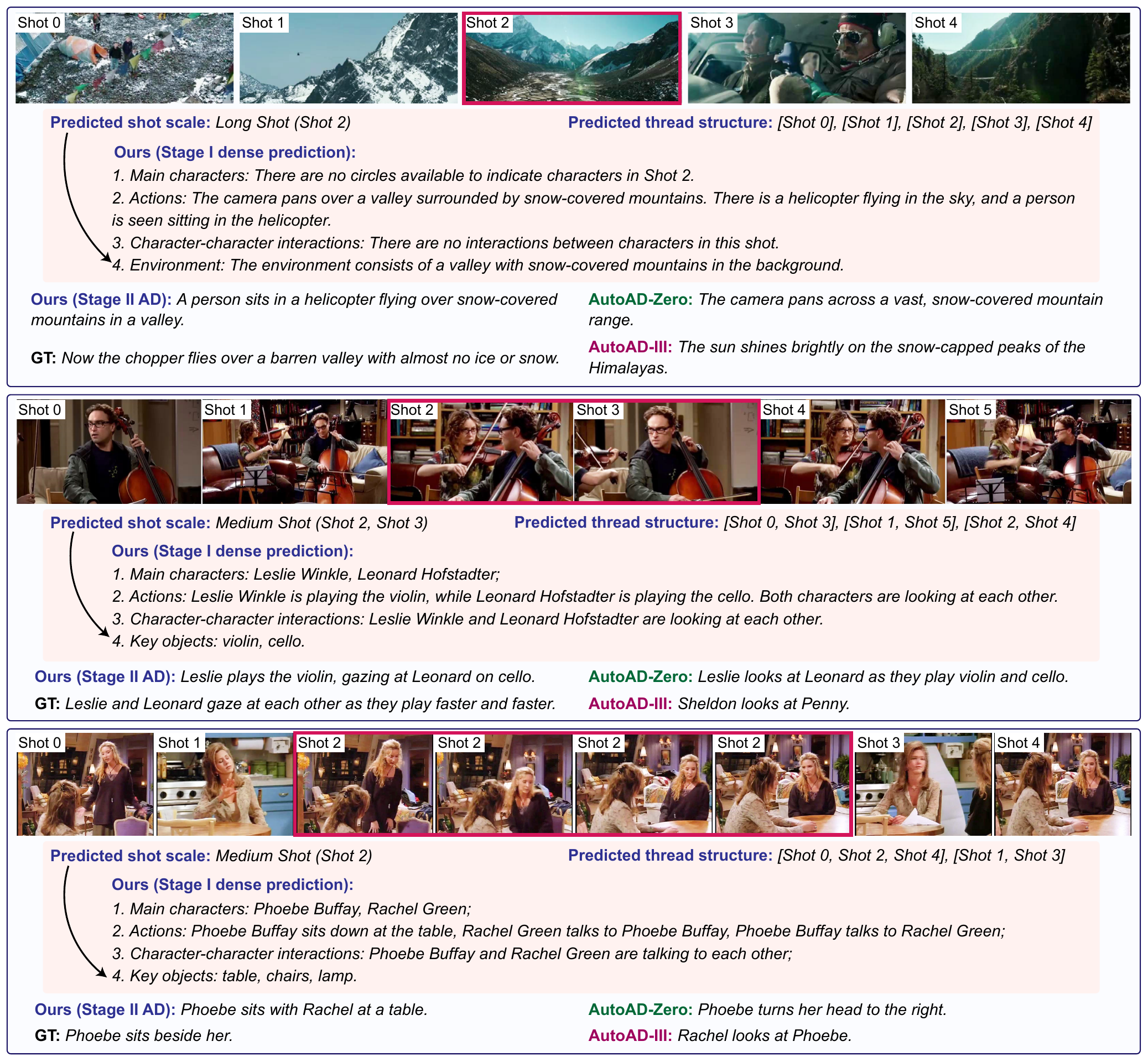}
\vspace{-0.2cm}
    \caption{
    \textbf{Detailed visualisations including intermediate results,} such as predicted thread structures and shot scales, as well as Stage I dense descriptions. Current shots (corresponding to AD intervals) are outlined by red boxes for illustration purposes only. Training-free methods adopt Qwen2-VL + LLaMA3 as base models. Examples from top to bottom are taken from \textit{Everest} (2015), \textit{The Big Bang Theory} (S1E5), and \textit{Friends} (S1E12), respectively.}
    \label{supfig:sup_vis_detail}
\end{figure*}

\vspace{3pt}
\noindent \textbf{Visualisation of intermediate outputs.}  
\cref{supfig:sup_vis_detail} provides more detailed visualisations, including intermediate results on thread structure and shot scale predictions, as well as Stage I descriptions. It also illustrates how the predicted shot scales influence the formulation of Stage I factors.

\begin{figure*}[htbp]
\centering
\includegraphics[width=0.99\textwidth]{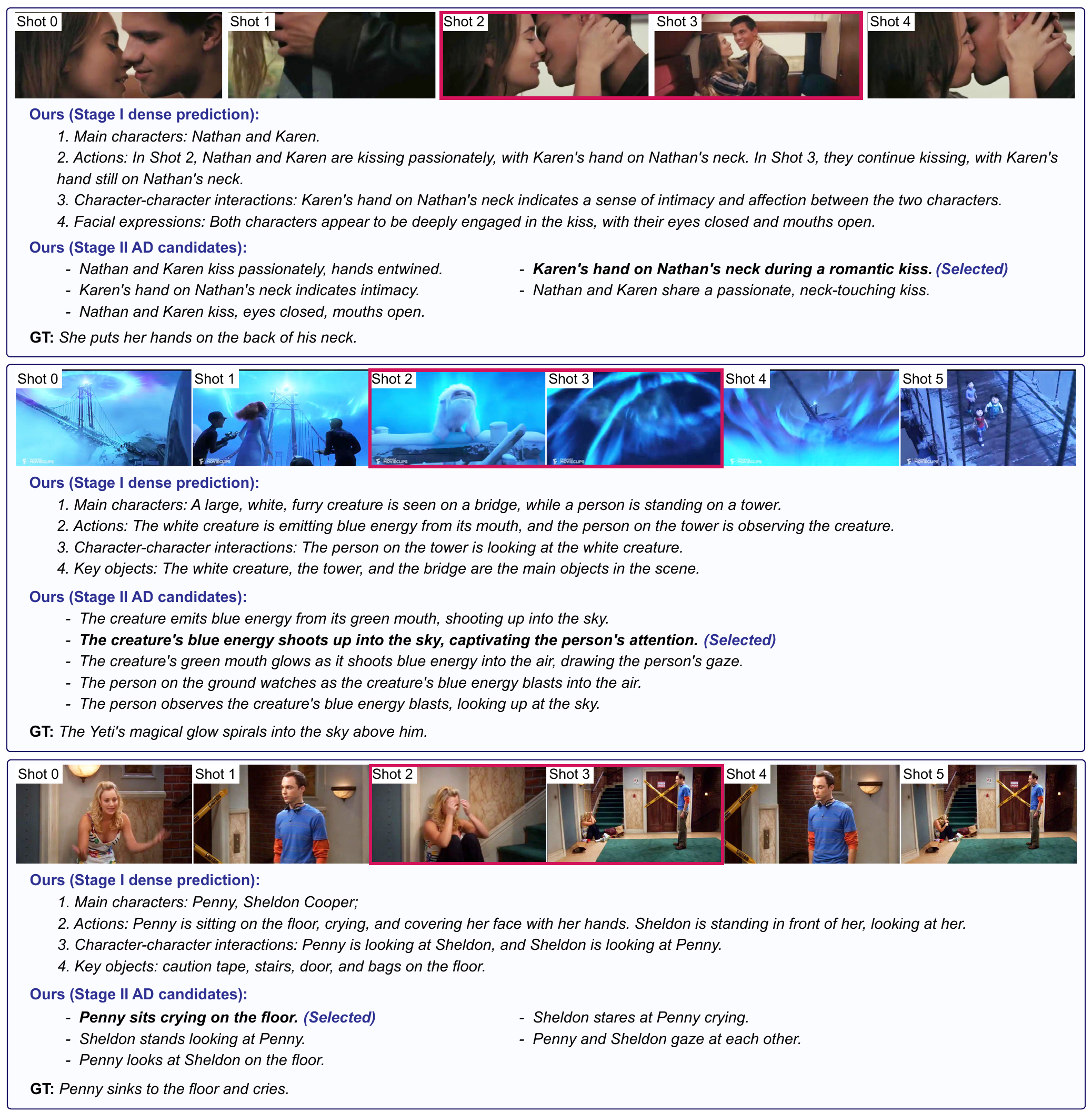}
\vspace{-0.2cm}
    \caption{
    \textbf{Visualisations for assisted AD generation,} where multiple AD candidates are extracted from the dense description, and the best among the five candidates is highlighted in \textbf{bold}. Current shots (corresponding to AD intervals) are outlined with red boxes for illustration purposes only. Our method uses Qwen2-VL and LLaMA3 as base models. The examples, from top to bottom, are taken from \textit{Abduction} (2011), \textit{Abominable} (2019), and \textit{The Big Bang Theory} (S2E3), respectively.
    }
    \label{supfig:sup_vis_multi}
\end{figure*}

\vspace{3pt}
\noindent \textbf{Visualisation for assisted AD generation.}  
\cref{supfig:sup_vis_multi} presents additional examples with Stage I descriptions, from which multiple AD candidates are extracted. Among the five AD predictions, the one that best aligns with the ground truth (based on the averaged CIDEr and action score) is highlighed.

In the top example, multiple actions are present in the Stage I dense description (e.g.\ \textit{``kiss''}, \textit{``hand on neck''}, \textit{``eyes closed''}, etc.), resulting in AD candidates that differ in both subjects and actions. In contrast, the middle example contains fewer actions (e.g.\ \textit{``shoot blue energy''} and \textit{``look''}/\textit{``observe''}). In this case, the AD candidates primarily vary in style, such as changes in subjects or sentence structures, providing varied options for selection.

\definecolor{DeepOrange}{rgb}{0.7, 0.3, 0}
\definecolor{DeepGreen}{rgb}{0.0, 0.5, 0.0}
\lstset{ 
  language=python, 
  basicstyle=\ttfamily\footnotesize, 
  keywordstyle=\color{blue}\bfseries,
  commentstyle=\color{DeepGreen},
  stringstyle=\color{DeepOrange},
  showstringspaces=false,
  numberstyle=\tiny\color{gray},
  numbersep=5pt,
  breaklines=true,
  breakindent=10pt,
  lineskip=1pt,
  captionpos=b,
  escapeinside={(*@}{@*)}
}
\begin{algorithm*}[htb!]
\caption{Stage I text prompt}\label{supalg:stage1}
\begin{lstlisting}
# Thread information injection
if exist(thread_structure):
    # {thread_structure}: e.g., "[Shot 1, Shot 3] share the same camera setup."
    thread_structure_text = "Finally, in one sentence, briefly explain why {thread_structure}\n"
else:
    thread_structure_text = ""

# Additional factors suggested by shot scales
factor_numbers = "four"
if effect_shot_scale <= 1.5:
    additional_factor_text = "4. Describe the facial expressions of characters.\n"
elif effect_shot_scale >= 2 and effect_shot_scale <= 3:
    additional_factor_text = "4. Describe the key objects that characters interact with.\n"
elif effect_shot_scale >= 3.5:
    additional_factor_text = "4. Describe the environment, focusing on the location, furniture, entrances and exits, etc.\n"
else:
    additional_factor_text = ""
    factor_numbers = "three"

# Stage I prompt
# {video_type}: "movie" or "TV series"
# {key_shots}: the middle shots (e.g., "[Shot 2, Shot 3]")
# {label_type}: "circles"
# {char_text}: character information (e.g., "Possible characters: Sheldon Cooper (red), ...")
prompt = (
    "Please watch the following {video_type} clip, where different shot numbers are labelled on the top-left of each frame.\n"
    f"Please briefly describe what happened in {key_shots} in the {factor_numbers} steps below:\n"
    f"1. Identify main characters (if {label_type} are available){char_text};\n"
    "2. Describe the actions of characters, i.e., who is doing what, focusing on the movements;\n" 
    "3. Describe the interactions between characters, such as looking;\n"
    f"{additional_factor_text}"
    f"Note, colored {label_type} are provided for character indications only, DO NOT mention them in the description. "   
    f"{thread_structure_text}"
    "Make sure you do not hallucinate information.\n"
    "### Answer Template ###\n" # Base format, need to be adjusted based on additionally factors, and whether the thread structure is injected
    "Description:\n"
    "1. Main characters: '';\n"
    "2. Actions: '';\n"
    "3. Character-character interactions: ''."
) 
\end{lstlisting}
\end{algorithm*}

\definecolor{DeepOrange}{rgb}{0.7, 0.3, 0}
\definecolor{DeepGreen}{rgb}{0.0, 0.5, 0.0}
\lstset{ 
  language=python, 
  basicstyle=\ttfamily\footnotesize, 
  keywordstyle=\color{blue}\bfseries,
  commentstyle=\color{DeepGreen},
  stringstyle=\color{DeepOrange},
  showstringspaces=false,
  numberstyle=\tiny\color{gray},
  numbersep=5pt,
  breaklines=true,
  breakindent=10pt,
  lineskip=1pt,
  captionpos=b,
  escapeinside={(*@}{@*)}
}
\begin{algorithm*}[htb!]
\caption{Stage II text prompt}\label{supalg:stage2}
\begin{lstlisting}
# Stage II system prompt
# {video_type}: "movie" or "TV series"
sys_prompt = (
    f"[INST] <<SYS>>\nYou are an intelligent chatbot designed for summarizing {video_type} audio descriptions. "
    "Here's how you can accomplish the task:------##INSTRUCTIONS: you should convert the predicted descriptions into one sentence. "
    "You should directly start the answer with the converted results WITHOUT providing ANY more sentences at the beginning or at the end. \n<</SYS>>\n\n{} [/INST] "
)

# Dataset dependent information
if dataset == "CMD-AD":
    verb_list = ['look', 'turn', 'take', 'hold', 'pull', 'walk', 'run', 'watch', 'stare', 'grab', 'fall', 'get', 'go', 'open', 'smile'] # top-15 lemma verb in the corresponding training set
    speed_factor = 0.275 # averaged (duration / number of words in AD) in the training set
elif dataset == "TV-AD":
    verb_list = ['look', 'walk', 'turn', 'stare', 'take', 'hold', 'smile', 'leave', 'pull', 'watch', 'open', 'go', 'step', 'get', 'enter']
    speed_factor = 0.2695
elif dataset == "MAD-Eval":
    verb_list = ['look', 'turn', 'sit', 'walk', 'take', 'stand', 'watch', 'hold', 'pull', 'see', 'go', 'open', 'smile', 'run', 'get']
    speed_factor = 0.5102

# Single AD generation / multiple AD candidate outputs (as an assistant)
if not assistant_mode: # Single AD
    pred_text = "Provide the AD from a narrator perspective.\n"
else: # Multiple ADs
    pred_text = "Provide 5 possible ADs from a narrator perspective, each offering a valid and distinct summary by emphasizing different key characters, actions, and movements present in the scene.\n"
    
# Stage II user prompt
# {text_pred}: Stage I dense description outputs
# {duration}: duration of the AD interval
# {example_sentence}: 10 randomly sampled AD sentences from training sets
user_prompt = (
    "Please summarize the following description for one movie clip into ONE succinct audio description (AD) sentence.\n"
    f"Description: {text_pred}\n\n"
    
    "Focus on the most attractive characters, their actions, and related key objects.\n"
    "For characters, use their first names, remove titles such as 'Mr.' and 'Dr.'. If names are not available, use pronouns such as 'He' and 'her', do not use expression such as 'a man'.\n"
    "For actions, avoid mentioning the camera, and do not focus on 'talking'.\n"
    "For objects, especially when no characters are involved, prioritize describing concrete and specific ones.\n"
    "Do not mention characters' mood.\n"
    "Do not hallucinate information that is not mentioned in the input.\n"
    f"Try to identify the following motions (with decreasing priorities): {verb_list}, and use them in the description.\n"
    "{pred_text}"
    f"Limit the length of the output within {int(duration / speed_factor + 1)} words.\n\n"
    
    "Output template (in JSON format): \"summarized_AD\": \"\".\n" # Adjust the template for single / multiple AD generation.
    "Here are some example outputs:\n"
    f"{example_sentence}"
)
\end{lstlisting}
\end{algorithm*}

\definecolor{DeepOrange}{rgb}{0.7, 0.3, 0}
\definecolor{DeepGreen}{rgb}{0.0, 0.5, 0.0}
\lstset{ 
  language=python, 
  basicstyle=\ttfamily\footnotesize, 
  keywordstyle=\color{blue}\bfseries,
  commentstyle=\color{DeepGreen},
  stringstyle=\color{DeepOrange},
  showstringspaces=false,
  numberstyle=\tiny\color{gray},
  numbersep=5pt,
  breaklines=true,
  breakindent=10pt,
  lineskip=1pt,
  captionpos=b,
  escapeinside={(*@}{@*)}
}
\begin{algorithm*}[htb!]
\caption{LLM-based character information removal text prompt}\label{supalg:char_info_removal}
\begin{lstlisting}
# System prompt for LLM-based character information removal in GT ADs
sys_prompt = (
    "You are an intelligent chatbot designed for removing character information of a sentence. "
    "Here's how you can accomplish the task: "
    "You should replace all character information including names, roles, and jobs into pronouns (e.g., he, she, they, her, him, them). "
    "Note, objects, locations, and animals are not counted as character information and should be kept as-is. "
    "You should output the result in JSON format WITHOUT providing ANY more sentences at the beginning or at the end."
)

# User prompt for LLM-based character information removal in GT ADs
# {text_gt}: GT AD
user_prompt = (
    "Please read the sentence below that describes a video clip:\n\n"
    f"Input sentence: \"{text_gt}\"\n\n"
    
    "Replace all character information including names, roles, and jobs into pronouns (e.g., he, she, they, her, him, them).\n"
    "Note, objects, locations, and animals are not counted as character information and should be kept as-is.\n"
    
    "**Examples:**\n"
    "   - Example 1:\n"
    "     - Input sentence: \"Spicoli watches Mr. Hand pass out the schedule.\"\n"
    "     - Ouput: \"He watches him pass out the schedule.\"\n"
    "   - Example 2:\n"
    "     - Input sentence: \"Waiting for a reply, the inspector has a look of smug satisfaction as he combs his neat moustache.\"\n"
    "     - Output: \"Waiting for a reply, he has a look of smug satisfaction as he combs his neat moustache.\"\n"
    "   - Example 3:\n"
    "     - Input sentence: \"Emmerich's eyebrows twitch as he watches her.\"\n"
    "     - Output: \"His eyebrows twitch as he watches her.\"\n"
    "   - Example 4:\n"
    "     - Input sentence: \"Inside is a second pair of doors.\"\n"
    "     - Output: \"Inside is a second pair of doors.\"\n"
    "   - Example 5:\n"
    "     - Input sentence: \"The blonde saunters over to him in her striped pantsuit and leans in for a kiss.\"\n"
    "     - Output: \"She saunters over to him in her striped pantsuit and leans in for a kiss.\"\n"
    "..." # More examples, omitted here for simplicity

    "**Output Format:**\n"
    "{\n"
    "  \"Output\": <output>\n"
    "}\n\n"
)
\end{lstlisting}
\end{algorithm*}

\definecolor{DeepOrange}{rgb}{0.7, 0.3, 0}
\definecolor{DeepGreen}{rgb}{0.0, 0.5, 0.0}
\lstset{ 
  language=python, 
  basicstyle=\ttfamily\footnotesize, 
  keywordstyle=\color{blue}\bfseries,
  commentstyle=\color{DeepGreen},
  stringstyle=\color{DeepOrange},
  showstringspaces=false,
  numberstyle=\tiny\color{gray},
  numbersep=5pt,
  breaklines=true,
  breakindent=10pt,
  lineskip=1pt,
  captionpos=b,
  escapeinside={(*@}{@*)}
}
\begin{algorithm*}[htb!]
\caption{LLM-based action sentence extraction text prompt}\label{supalg:action}
\begin{lstlisting}
# System prompt for LLM-based action sentence extraction from GT ADs
sys_prompt = (
    "You are an intelligent chatbot designed for decompose the sentence into subsentences. "
    "Here's how you can accomplish the task: "
    "You should split (rewrite if needed) the sentence into subsentences, each containing only one action phrase (i.e., verb phrase). "
    "You should output your answer in JSON format WITHOUT providing ANY more sentences at the beginning or at the end."
)

# User prompt for LLM-based action sentence extraction from GT ADs
# {text_gt}: GT AD after character information removal
user_prompt = (
    "Please read the sentence below that describes a video clip:\n\n"
    f"Input sentence: \"{text_gt}\"\n\n"
    "Split and rewrite the sentence into subsentences, each containing only one action (i.e., verb phrase) and preserving all other information (e.g., locations, time, affections, etc.).\n"
    "Do not output repeating actions.\n"
    "**Examples:**\n"
    "   - Example 1:\n"
    "     - Input sentence: \"He watches him pass out the schedule.\"\n"
    "     - Subsentences: [\"He watches him.\", \"He passes out the schedule.\"]\n"
    "   - Example 2:\n"
    "     - Input sentence: \"Waiting for a reply, he has a look of smug satisfaction as he combs his neat moustache.\"\n"
    "     - Subsentences: [\"He waits for a reply.\", \"He has a look of smug satisfaction.\", \"He combs his neat moustache.\"]\n"
    "   - Example 3:\n"
    "     - Input sentence: \"He swings in front of Kingpin, then bounces off a building and kicks the criminal into the air.\"\n"
    "     - Subsentences: [\"He swings in front of him.\", \"He bounces off a building.\", \"He kicks him into the air.\"]\n"
    "   - Example 4:\n"
    "     - Input sentence: \"His eyebrows twitch as he watches her.\"\n"
    "     - Subsentences: [\"His eyebrows twitch.\", \"He watches her.\"]\n"
    "   - Example 5:\n"
    "     - Input sentence: \"Inside is a second pair of doors.\"\n"
    "     - Subsentences: [\"Inside is a second pair of doors.\"]\n"
    "..." # More examples, omitted here for simplicity
    
    "**Output Format:**\n"
    "{\n"
    "  \"Subsentences\": \n"
    "  [\n"
    "    <subsentence1>,\n"
    "    <subsentence2>,\n"
    "    <subsentence3>,\n"
    "    ...\n"
    "  ]\n"
    "}\n\n"
)
\end{lstlisting}
\end{algorithm*}

\definecolor{DeepOrange}{rgb}{0.7, 0.3, 0}
\definecolor{DeepGreen}{rgb}{0.0, 0.5, 0.0}
\lstset{ 
  language=python, 
  basicstyle=\ttfamily\footnotesize, 
  keywordstyle=\color{blue}\bfseries,
  commentstyle=\color{DeepGreen},
  stringstyle=\color{DeepOrange},
  showstringspaces=false,
  numberstyle=\tiny\color{gray},
  numbersep=5pt,
  breaklines=true,
  breakindent=10pt,
  lineskip=1pt,
  captionpos=b,
  escapeinside={(*@}{@*)}
}
\begin{algorithm*}[htb!]
\caption{LLM-based action metric text prompt}\label{supalg:llm_based}
\begin{lstlisting}
# System prompt for LLM-based action evaluation
sys_prompt = (
    "You are an evaluation assistant designed to assess the accuracy of a description (Des) in capturing the action specified in a reference sentence (Ref) for a movie clip. "
    "Focus only on the presence of the referenced action and ignore any additional, unrelated actions in the description. "
    "Ignore any character information in the description. "
    "Avoid assumptions about action details beyond what is explicitly provided in either the reference or description. "
    "Output the result exclusively in JSON format, with a score (0 to 3) and a brief explanation describing the relationship between the actions in Ref and Des, without any introductory or concluding text."
)

# User prompt for LLM-based action evaluation
# {text_gt}: character-free action sentence extracted from GT AD
# {text_pred}: predicted dense description (paragraph) or AD sentence
user_prompt = (
    "You will be provided with a reference action sentence (Ref) and a description (Des) for a clip. "
    "Your task is to evaluate if the action described in Ref is explicitly or clearly implied in Des. "
    "Focus only on the presence of the referenced action, without considering any additional actions and character information that may appear in Des, and do not assume any actions beyond those stated in Ref or Des. "
    "The output should be a score (0 to 3) with a brief one-sentence explanation describing the relationship between the actions in Ref and Des.\n\n"

    "# Scoring Criteria:\n"
    "- **Score 0:** The action in Ref is completely unrelated to actions in Des.\n"
    "- **Score 1:** The action in Ref is loosely related to an action in Des.\n"
    "- **Score 2:** The action in Ref is similar in meaning to an action in Des.\n"
    "- **Score 3:** The action in Ref exactly matches an action in Des, using the same verb.\n\n"
    "# Examples:\n"
    "- Example 1:\n"
    "  - Ref: 'He runs across the street.'\n"
    "  - Des: 'Tom is jogging down the street.'\n"
    "  - Output: {\n"
    "      'score': 2,\n"
    "      'explanation': 'The action \"runs across the street\" in Ref is similar to \"jogging down the street\" in Des.'\n"
    "    }\n\n"  
    "- Example 2:\n"
    "  - Ref: 'He pours wine into a glass.'\n"
    "  - Des: 'The woman drinks.'\n"
    "  - Output: {\n"
    "      'score': 1,\n"
    "      'explanation': 'The action \"pours wine into a glass\" in Ref is loosely related to \"drinks\" in Des.'\n"
    "    }\n\n"
    "..." # More examples, omitted here for simplicity
    "# Output Format:\n"
    "{\n"
    "  'score': <score>,\n"
    "  'explanation': '<explanation>'\n"
    "}\n\n"
    "# Now, apply these instructions to the following texts:\n\n"
    f"   - # Reference (Ref): '{text_gt}'\n"
    f"   - # Description (Des): '{text_pred}'"
)
\end{lstlisting}
\end{algorithm*}

\end{document}